\documentclass[]{elsarticle}

\usepackage[hyphens]{url}
\usepackage[hidelinks]{hyperref}
\usepackage{listings}
\usepackage[usenames,dvipsnames]{color}
\usepackage{csquotes}
\usepackage{dsfont}
\usepackage{caption}
\usepackage{subfig}
\usepackage{microtype}

\usepackage{booktabs}
\usepackage{pbox}
\usepackage{soul}
\usepackage[final]{changes}

\biboptions{numbers,sort&compress}

\usepackage{tikz}
\usetikzlibrary{shapes.geometric}
\usetikzlibrary{positioning,shapes,shadows,arrows,calc}
\usetikzlibrary{intersections}
\pgfdeclarelayer{background}
\pgfdeclarelayer{foreground}
\pgfsetlayers{background,main,foreground}

\newcommand{\emcite}[1]{\citeauthor{#1} (\citeyear{#1})}

\newcommand{\hh}[1]{#1}
\newcommand{\fh}[1]{#1}
\newcommand{\pk}[1]{#1}

\usepackage{amsmath}
\usepackage{amsthm}
\usepackage{amssymb}
\newtheorem{thm}{Theorem}

\newtheorem{define}[thm]{Definition}

\newcommand{\system}[1]{\textit{#1}}

\newcommand{\changed}[1]{{#1}}

\newcommand{\changedTwo}[1]{{#1}}

\newcommand{\changedThree}[1]{{#1}}

\DeclareMathOperator*{\argmax}{arg\,max}

\usepackage{todonotes}

\begin{document}

\title{ASlib: A Benchmark Library for Algorithm Selection}


\address[muc]{LMU Munich, Germany}
\address[muenster]{University of M\"unster, Germany}
\address[freiburg]{University of Freiburg, Germany}
\address[vancouver]{University of British Columbia, Vancouver, Canada}
\address[paderborn]{University of Paderborn, Germany}
\address[tue]{Eindhoven Institute of Technology, Netherlands}
\address[ibm]{IBM Research, United States}

\author[muc]{Bernd Bischl}
\ead{bernd.bischl@stat.uni-muenchen.de}

\author[muenster]{Pascal Kerschke}
\ead{kerschke@uni-muenster.de}

\author[vancouver]{Lars Kotthoff}
\ead{larsko@cs.ubc.ca}

\author[freiburg]{Marius Lindauer}
\ead{lindauer@cs.uni-freiburg.de}

\author[ibm]{\mbox{Yuri Malitsky}}
\ead{yuri.malitsky@gmail.com}


\author[vancouver]{Alexandre Fr\'{e}chette} 
\ead{afrechet@cs.ubc.ca}

\author[vancouver]{Holger Hoos} 
\ead{hoos@cs.ubc.ca}

\author[freiburg]{Frank Hutter} 
\ead{fh@cs.uni-freiburg.de}

\author[vancouver]{\mbox{Kevin Leyton-Brown}} 
\ead{kevinlb@cs.ubc.ca}

\author[paderborn]{Kevin Tierney} 
\ead{tierney@dsor.de}

\author[tue]{Joaquin Vanschoren} 
\ead{j.vanschoren@tue.nl}

\begin{abstract}
The task of algorithm selection involves choosing an algorithm from a set of
algorithms on a per-instance basis in order to exploit the varying performance
of algorithms over a set of instances. The algorithm selection problem is
attracting increasing attention from researchers and practitioners in AI.
Years of fruitful applications in a number of domains have resulted in a large
amount of data, but the community lacks a standard format or repository for
this data. This situation makes it difficult to share and compare different
approaches effectively, as is done in other, more established fields. It also
unnecessarily hinders new researchers who want to work in this area.
To address this problem, we introduce a standardized format for representing algorithm selection scenarios
and a repository that contains a growing number of data sets from the
literature. Our format has been designed to be able to express a wide variety of
different scenarios. \changedTwo{To demonstrate} the breadth and power of our platform, we describe a study
that builds and evaluates algorithm selection models through a common
interface. The results display the potential of algorithm selection to achieve significant performance improvements \hh{across a broad range of problems and algorithms.}
\end{abstract}


\maketitle


\section{Introduction}
\label{sec:intro}


Although NP-complete problems are widely believed to be intractable in the worst case, it is often possible to solve even very large instances of such problems that arise in practice. This is fortunate, because such problems are ubiquitous in Artificial Intelligence applications. There has thus emerged a large subfield of AI devoted to the advancement and analysis of heuristic algorithms for attacking hard computational problems. Indeed, quite surprisingly, this subfield has made consistent and substantial progress over the past few decades, with the newest algorithms quickly solving benchmark problems \changedTwo{that were beyond reach until recently}. The results of the international SAT competitions provide a paradigmatic example of this phenomenon. Indeed, the importance of this competition series has gone far beyond documenting the \hh{progress achieved by the SAT community in solving difficult and application-relevant SAT instances}---it has been instrumental in driving research itself, helping the community to coalesce around a shared set of benchmark instances and providing an impartial basis for determining which new ideas yield the biggest performance gains.

The central premise of events like the SAT competitions is that the research community ought to build, identify and reward single solvers that achieve strong across-the-board performance. However, this quest appears quixotic: 
most hard computational problems admit multiple solution approaches, none of which dominates all alternatives across multiple problem instances. In particular, this fact has been observed to hold across a wide variety of AI applications, including
propositional satisfiability (SAT)~\cite{xu2012evaluating}, constraint
satisfaction (CSP)~\cite{cphydra}, planning~\cite{howe_exploiting_1999,helmert_fast_2011}, and
supervised machine
learning~\cite{ThoHutHooLey13-AutoWEKA,Vanschoren2012,feurer-nips15a}. 
An alternative is to accept that no single algorithm will offer the best performance on all instances, and instead aim to identify a portfolio of complementary algorithms and a strategy for choosing between them \citep{rice_algorithm_1976}. To see the appeal of this idea, consider the results of the sequential application (SAT+UNSAT) track of the $2014$ SAT Competition.\footnote{\url{http://www.satcompetition.org/2014/results.shtml}} The best of the $35$ submitted solvers, \texttt{Lingeling ayv}~\cite{lingeling}, solved $77\%$ of the $300$ instances.
However, if we could somehow choose the best among these $35$ solvers on a per-instance basis, we would be able to solve $92\%$ of the instances. 

\fh{Research on this \emph{algorithm selection problem}~\citep{rice_algorithm_1976} has demonstrated the practical feasibility of using machine learning for this task.}
In fact, although practical algorithm selectors occasionally \fh{choose suboptimal algorithms}, \hh{their performance can 
	get close to that of an oracle that always makes the best choice.}
The area \hh{began to attract} considerable attention when methods based on algorithm selection began to outperform standalone solvers in SAT competitions~\cite{xu_satzilla_2008}. Algorithm selectors have since come to dominate the state of the art on many other problems, including
CSP~\cite{cphydra}, planning~\cite{helmert_fast_2011}, Max-SAT~\cite{malitsky_evolving_2013}, QBF~\cite{pulina_self-adaptive_2009}, and ASP~\cite{gebser_portfolio_2011}.

To date, much of the progress in research on algorithm selection has been
demonstrated in algorithm competitions originally intended for
non-portfolio-based (``standalone'') solvers. This has given rise to a variety
of challenges for the field. First, benchmarks selected for such competitions tend
to emphasize problem instances that are currently hard for existing standalone
algorithms (to drive new research on solving strategies) rather than the wide
range of both easy and hard instances that would be encountered in practice
(which would be appropriately targeted by researchers developing algorithm
selectors). Relatedly, benchmark sets change from year to year, making it
difficult to assess the progress of algorithm selectors over time. Second,
although competitions often require entrants to publish their source code, none
require entries based on algorithm selectors to publish the code used to
\emph{construct} the algorithm selector (e.g., via training a machine learning
model) or to adhere to a consistent input format. Third, overwhelming competition
victories by algorithm selectors can make it more difficult for new standalone
solver designs to get the attention they deserve and can thus create resentment
among solver authors. Such concerns have led to a backlash against the
participation of portfolio-based solvers in competitions; for example,
\hh{starting in 2013 solvers that explicitly combine more than two component
algorithms have been excluded from the SAT competitions}.
\changedTwo{For similar reasons, there is a specific prize for non-portfolio solvers
in the learning track of the International Planning Competition~\cite{VallatiCGMRS15}.}

The natural solution to these challenges is to evaluate algorithm selectors on
their own terms rather than trying to shoehorn them into competitions intended
for standalone solvers. This article, written by a large set of authors active in research
on algorithm selectors, aims to advance this goal by introducing a set of
specifications and tools designed to standardize and facilitate such evaluations. Specifically, we propose a benchmark library, called ASlib, tailored to the cross-domain evaluation of algorithm selection techniques. 
%
\changedTwo{In Section~\ref{sec:spec}, we provide a summary of the data format specification used in ASlib}
that covers a wide variety of foreseeable evaluations. To date, we have instantiated this specification with
benchmarks from six different problem domains, which we describe in Section~\ref{sec:scenarios}.
However, we intend for ASlib to grow and evolve over time.
Thus, our article is accompanied by an online repository
(\url{http://aslib.net}), which accepts submissions from any researcher. Indeed,
we \changedTwo{already} included scenarios that have been submitted by \changedTwo{contributors outside the core group of ASlib maintainers}.

Our system automatically checks newly submitted datasets to verify that they adhere to the
specifications and then provides an overview of the data, including the results
of some straightforward algorithm selection approaches based on regression, clustering and classification.
We provide some examples of these automatically-generated overviews and benchmark results in Sections \ref{sec:eda} and \ref{sec:experiments}.
All code used to parse the format files, explore the algorithm selection scenarios and run benchmark machine
learning models on them is publicly available in a new R package dubbed
\system{aslib}.\footnote{This package is currently hosted at
	\url{https://github.com/coseal/aslib-r}. We will submit it to the official R package server CRAN alongside the final version of this article.}
\changedTwo{In Section~\ref{sec:comp}, we discuss two recent examples of competition settings using ASlib, along with their advantages and disadvantages.}

Overall, our main objective in creating ASlib is the same as that of an algorithm competition: to allow
researchers to compare their algorithms systematically and fairly, without having to replicate someone else's system or to personally collect
raw data. We hope that it will help the community to obtain an unbiased understanding of
the strengths and weaknesses of different methodologies and thus to improve the current state of the art in per-instance algorithm selection.

%


\section{Background}
\label{sec:related_work}

\citet{rice_algorithm_1976} was the first to formalize the idea of selecting among different algorithms on a per-instance basis. While he referred to the problem simply as \emph{algorithm selection}, we prefer the more precise term \emph{per-instance algorithm selection}, to avoid confusion with the (simpler) task of selecting one of several given algorithms to optimize performance on a given set or distribution of instances.

\begin{define}[Per-instance algorithm selection problem]
	\label{def:algo_sel}
	Given
	\begin{itemize}
	  \item \changedTwo{a set $\mathcal{I}$ of problem instances drawn from a distribution $\mathcal{D}$},
	  \item a space of algorithms $\mathcal{A}$, and
	  \item a performance measure $m: \mathcal{I} \times \mathcal{A} \rightarrow \mathds{R}$,
	\end{itemize}
	the \emph{per-instance algorithm selection problem}	is to 
	find a mapping
        $s: \mathcal{I} \rightarrow \mathcal{A}$ that optimizes $\mathds{E}_{i \sim \mathcal{D}}m(i, s(i))$,
	i.e., the \changedTwo{expected performance measure for instances $i$ distributed according to $\mathcal{D}$, achieved by running the selected algorithm $s(i)$ for instance $i$}.
\end{define}


In practice, the mapping $s$ is often implemented by using 
so-called instance features, i.e., 
characterizations of the instances $i \in \mathcal{I}$.
These instance features are then mapped to an algorithm using 
machine learning techniques. However, the computation of instance features incurs additional costs,
which have to be considered in the performance measure $m$.



There are many ways of tackling per-instance algorithm selection and related
problems. Almost all contemporary approaches use machine learning 
to build predictors of the behaviour of given algorithms as a function of instance features. 
This general strategy may involve a single learned model or a complex combination of several, which, given a new problem instance 
to solve, is used to decide which algorithm or which combination of
algorithms to choose.

\subsection{What to select and when}

It is perhaps most natural to select a single
algorithm for solving a given problem instance. This \changedTwo{approach is, e.g., used in the
SATzilla~\cite{Satzilla03,xu_satzilla_2008},
\textsc{ArgoSmArT}~\cite{nikoli_instance-based_2009},
SALSA~\cite{demmel_self-adapting_2005} and
\textsc{Eureka}~\cite{cook_maximizing_1997} systems. Its main disadvantage}
is that there is no way of mitigating a
poor selection---\changedTwo{the system cannot recover if the algorithm it chose for a problem instance exhibits poor performance.}

Alternatively, we can seek a schedule that determines an ordering and time budget according to which we run all or a subset of the algorithms in the portfolio; usually, this schedule 
is chosen in a way that reflects the expected performance of the given
algorithms (see, e.g.,
\cite{pulina_self-adaptive_2009,cphydra,kadioglu_algorithm_2011,hoos_aspeed_2014,howe_exploiting_1999}).
Under some of these approaches, the computation of the schedule is
treated as an optimization problem that aims to maximize, e.g., the number of
problem instances solved within a timeout. For stochastic algorithms, the further
question of whether and when to restart an algorithm arises, opening the possibility of schedules that contain only a single algorithm, restarted several times (see, e.g., \cite{gomes_algorithm_2001,cicirello_max_2005,streeter_restart_2007,gagliolo_learning_restart_strategies_2007}).
Instead of performing algorithm selection only once before starting to solve a problem, selection can
also be carried out repeatedly while the instance is being solved, taking into account information revealed during the algorithm run. 
Such methods monitor the execution of
the chosen algorithm(s) and take remedial action if performance deviates from
what is expected~\cite{gagliolo_adaptive_2004,MUSportfolio,LeiteBV12}, or perform selection 
repeatedly for subproblems of the given instance \cite{LAG1,LAG2,arbelaez_continuous_2010,samulowitz_learning_2007}.

\subsection{How to select}\label{sec:background:how}

The kinds of decisions the selection process is asked to produce drive the choice of machine learning models that perform the selection. If only a
single algorithm should be run, we can train a classification model that makes
exactly that prediction. This renders algorithm selection conceptually quite
simple---only a single machine learning model needs to be trained and run to
determine which algorithm to choose (see, e.g., \cite{guerri_learning_2004,gent_learning_2010,malitsky_non-model-based_2011}).

There are alternatives to using a classification model to select a single algorithm to be run on
a given instance, such as using regression models to predict the performance of each algorithm in the
portfolio. This regression approach
was adopted by several systems~\cite{Satzilla03,xu_satzilla_2008,roberts_learned_2007,silverthorn_latent_2010,Mersmann2013}.
Other approaches include 
the use of clustering techniques to partition problem instances in feature space and make decisions for each partition
separately~\cite{stergiou_heuristics_2009,kadioglu_isac_2010}, hierarchical models that make a series of
decisions~\cite{xu_hierarchical_2007,hurley_proteus_2014}, cost-sensitive support vector
machines~\cite{Bischl2012_2} and cost-sensitive decision forests~\cite{xu_hydra-mip_2011}.

\subsection{Selection enablers}


In order to make their decisions, algorithm selection systems need information
about the problem instance to solve and the performance of the algorithms in the
given portfolio. The extraction of this information---the
features used by the machine learning techniques used for selection---incurs overhead not required
when only a single algorithm is used for all instances regardless of instance characteristics. 
It is therefore desirable
to extract information as cheaply as possible, thus ensuring that the performance benefits
of using algorithm selection are not outweighed by this overhead.

Some approaches use only past performance of the algorithms in the portfolio
as a basis for selecting the one(s) to be run on a given problem
instance~\cite{gagliolo_adaptive_2004,streeter_combining_2007,silverthorn_latent_2010}. This approach has the benefit that the required data can be collected with minimal overhead as algorithms are executed. 
It can work well if the performance of the algorithms is similar
on broad ranges of problem instances. However, when this assumption is not 
satisfied (as is often the case), more informative features are needed.


Turning to richer instance-specific features, commonly used features include the number of variables of a problem instance and
properties of the variable domains (e.g., the list of possible assignments in
constraint problems, the number of clauses in SAT, the number of goals in
planning). 
Deeper analysis can involve properties of graph representations derived from the
input instance (such as the constraint graph~\cite{leyton2003portfolio,gent_learning_2010}) or properties of encodings into
different problems (such as SAT features for SAT-encoded planning
problems~\cite{fawcett2014improved}).

In addition, features can be extracted from
short runs of one or more solvers on the given problem instance. 
Examples of such probing features include the number of search
nodes explored within a certain time, the fraction of partial solutions that are
disallowed by a certain constraint or clause, the average depth reached
before backtracking is required, or characteristics of local minima found quickly using local search.
Probing features are usually more expensive to
compute than the features that can be obtained from shallow analysis of the instance
specification, \fh{but they can also be more powerful and have thus been used by many authors (see, e.g., 
\citep{nudelman_understanding_2004,cphydra,pulina_multi-engine_2007,xu_satzilla_2008,hutter2014algorithm}).}
%
For continuous blackbox optimization, algorithm selection can be performed based on Exploratory Landscape Analysis
\cite{Mersmann2013,Bischl2012_2,Kerschke2014}. 
The approach defines a set of numerical features (of different complexities and computational costs) 
to describe the landscapes of such optimization problems. Examples range from simple features that 
describe the distribution of sampled objective values to more expensive probing features based on local search. 

\changed{Finally, in the \fh{area of meta-learning (learning about the performance of machine learning algorithms; for an overview, see, e.g, \citep{Brazdil_metalearning_2008}), these} features are known as \emph{meta-features}. They include statistical and information-theoretical measures (e.g., variable entropy), landmarkers (measurements of the performance of fast algorithms~\cite{Pfahringer:2000p553}), sampling landmarkers (similar to probing features) and model-based meta-features~\cite{Vanschoren2010}. These meta-features, and the past performance measurements of many machine learning algorithms, are available from the online machine learning platform OpenML~\cite{openml2013}. In contrast to ASlib, however, OpenML is not designed to allow cross-domain evaluation of algorithm selection techniques.
}


\subsection{Algorithm Selection and Algorithm Configuration}

A problem closely related to algorithm selection is the algorithm configuration problem:
given a parameterized algorithm $A$, a set of problem instances $I$ and a performance measure $m$, find a parameter setting of $A$ that optimizes $m$ on $I$ \changedTwo{(see \cite{hutter_paramils_2009} for a formal definition)}.
While algorithm selection operates on finite (usually small) sets of algorithms, algorithm configuration operates on the combinatorial space of an algorithm's parameter settings. General algorithm configuration methods, such as ParamILS~\cite{hutter_paramils_2009}, GGA~\cite{ansotegui_gender-based_2009}, I/F-Race~\cite{BirEtAl10}, and SMAC~\cite{HutHooLey11-SMAC}, 
have yielded substantial performance improvements (sometimes orders of magnitude
speedups) of state-of-the-art algorithms for several benchmarks, including
SAT-based formal verification~\cite{HutBabHooHu07}, mixed integer
programming~\cite{HutHooLey10-mipconfig}, AI planning~\cite{roberts_what_2008,Vallati13-SOCS}, the combined selection and hyperparameter optimization of machine learning algorithms~\cite{ThoHutHooLey13-AutoWEKA}, \changedTwo{and joint architecture and hyperparameter search in deep learning~\cite{domhan-ijcai15a}}.
Algorithm configuration and selection are complementary since configuration can
identify algorithms with peak performance for homogeneous benchmarks and
selection can then choose from among these specialized algorithms. Consequently, several possibilities exist for combining algorithm configuration and selection~\cite{hutter_performance_2006,xu_hydra_2010,kadioglu_isac_2010,xu_hydra-mip_2011,malitsky_evolving_2013,sabharwal_boosting_2013,ISAC++,feurer-aaai15a}.
The algorithm configuration counterpart of ASlib is AClib~\cite{HutEtAl14:AClib} (\url{http://aclib.net}).
\changed{In contrast to ASlib, it is infeasible in AClib to store performance data for all possible parameter configurations, which often number more than $10^{50}$.
Therefore, an experiment on AClib includes new (expensive) runs of the target algorithms with different configurations
and hence, experiments on AClib are a lot more costly than experiments on ASlib, where no new algorithm runs are necessary.\footnote{\fh{In algorithm configuration, this need for expensive runs indeed causes a problem for research. One way of mitigating it is offered by fast-to-evaluate surrogate algorithm configuration benchmarks~\cite{Eggensperger2015}.}}} \changedTwo{Furthermore, in contrast to AClib, ASlib does not include the actual instances and binaries of the algorithms.
Therefore, ASlib does not provide a way to generate new performance data, as is required in AClib as a consequence of the need to assess the performance of new target algorithm configurations arising within the configuration process. 
However, ASlib and AClib can be combined by generating actual performance data based on the resources in AClib and then creating an ASlib scenario which selects between different solver configurations on a per-instance basis.
}




A full coverage of the wide literature on algorithm selection is beyond the scope of this article, but we refer the interested reader to recent survey articles on the topic~\cite{smith-miles_cross-disciplinary_2009,kotthoff_algorithm_2014,Serban:2013,VallatiCK15}.

\section{Summary of Format Specification}
\label{sec:spec}

We propose a data format specification for algorithm selection scenarios,
i.e., instances of the per-instance algorithm selection problem.
This format and the resulting data repository allow a fair and convenient 
scientific evaluation and comparison of algorithm selectors.

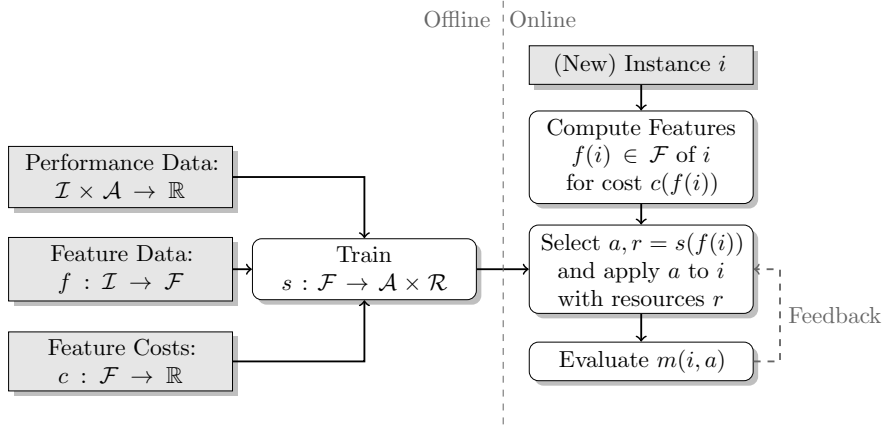
\begin{figure}[tbp]
\resizebox{\textwidth}{!}{\tikzstyle{activity}=[rectangle, draw=black, rounded corners, text centered, text width=9em, fill=white, drop shadow]
\tikzstyle{data}=[rectangle, draw=black, text centered, fill=black!10, text width=9em, drop shadow]
\tikzstyle{myarrow}=[->, thick]
\begin{tikzpicture}[align=center,node distance=3.7cm]
	
	\node (perf) [data] {Performance Data: $\mathcal{I} \times \mathcal{A} \to \mathbb{R}$};
	\node (feat) [data, below of=perf, yshift=2.3cm] {Feature Data: $f: \mathcal{I} \to \mathcal{F}$};
	\node (cost) [data, below of=feat, yshift=2.3cm] {Feature Costs: $c: \mathcal{F} \to \mathbb{R}$};
	
	\node (train) [activity, right of=feat] {Train\\ $s: \mathcal{F} \to \mathcal{A} \times \mathcal{R}$};
	
	\node (run) [activity, right of=train, xshift=0.5cm] {Select $a, r = s(f(i))$ and apply $a$ to $i$ with resources $r$};
	\node (feat_i) [activity, above of=run, yshift=-2.0cm] {Compute Features $f(i) \in \mathcal{F}$ of $i$ for cost $c(f(i))$};
	\node (inst) [data, above of=feat_i, yshift=-2.3cm] {(New) Instance $i$};
	
	\node (eva) [activity, below of=run, yshift=2.3cm] {Evaluate $m(i, a)$};
	
	\draw[myarrow] (perf) -| (train.north);
	\draw[myarrow] (feat) -- (train.west);
	\draw[myarrow] (cost) -| (train.south);
	
	\draw[myarrow] (inst) -- (feat_i);
	\draw[myarrow] (feat_i) -- (run);
	\draw[myarrow] (train) -- (run);
	\draw[myarrow] (run) -- (eva);
	\draw[myarrow, dashed, color=black!60] (eva.east) -- ($(eva.east)+(0.4,0.0)$) |- node [right, yshift=-0.7cm, color=black!60] {Feedback} (run.east);
	
	\path (train.east -| train.east)+(0.4,4.2) node (a) {};
    \path (train.east |- train.east)+(0.4,-2.5) node(b) {};
    \draw[dashed, color=black!60] (a) -- (b);
    \path[color=black!60] (train.east -| train.east)+(-0.3,3.8) node (a) {Offline};
    \path[color=black!60] (train.east -| train.east)+(1.0,3.8) node (a) {Online};
	
\end{tikzpicture}}
\caption{Algorithm Selection workflow.}
\label{dia:as}
\end{figure}

The format specification assumes a generic approach to algorithm selection,
depicted in Figure~\ref{dia:as}. The general approach is as follows.



\begin{enumerate}
  \item \changed{A vector of instance features $f(i) \in \mathcal{F}$} of $i$ is computed. 
  		Feature computation may occur in several stages,
  		each of which produces \changed{a group of (one or more) features}. Furthermore, later stages may depend on the results of earlier ones.
  	    Each feature group incurs a cost,
  	    e.g., runtime. If no features are required, the cost is $0$ (this occurs, e.g., for variants of algorithm selection that compute static schedules).
  \item A machine learning technique $s$ selects an algorithm $a \in \mathcal{A}$ based on the feature vector from Step 1.
  \item The selected algorithm $a$ is applied to $i$.
  \item Performance measure $m$ is evaluated, taking into account feature computation costs and the performance of the selected algorithm.
  \item \changed{Some algorithm selectors do not select a single algorithm, but
  compute a schedule of several algorithms:
  		they apply $a$ to $i$ for a resource budget $r \in \mathcal{R}$
        (e.g., CPU time), evaluate the performance metric, evaluate
        a stopping criterion, and repeat as necessary, taking observations made
        during the run of $a$ into account.\footnote{In \fh{principle}, the workflow can be arbitrarily more complex, e.g., alternating between computing further features and running selected algorithms.}}
\end{enumerate}

\changed{The purpose of our library is to provide all information necessary for performing algorithm selection experiments
using the given scenario data.
The user does not need to actually run algorithms on instances, as all performance data is already \fh{precomputed}. 
\hh{This drastically reduces the time required for executing studies, i.e., 
the runtime of studies is now} dominated by the time required for learning $s$
and not by applying algorithms to instances (e.g., solving SAT problems).
It also means that results are perfectly \hh{reproducible};
for example, the runtimes of algorithms
do not depend on the hardware used;
\hh{rather, they can be simply looked up in the performance data for a} scenario.
}

Table~\ref{nutshell:spec} shows the basic structure of a scenario definition in
ASlib;
the complete specification with all details can be found in \fh{an accompanying technical report}~\cite{spec} and on our online platform.



\noindent\fbox{
\begin{minipage}[tbh]{0.96\textwidth} 
\captionof{table}{Overview of a scenario in the ASlib format.}
\label{nutshell:spec}
\paragraph{Mandatory Data}
\begin{itemize}
  	\item The \textit{meta information file} is a global description file containing general information about 
		the scenario, including the name of the scenario, performance measures, algorithms, features and 
		limitations of computational resources.
	\item The \textit{algorithm performance} file contains performance measurements \changedTwo{with possible repetitions} and completion status of the algorithm runs. \changedTwo{The performance metric can be arbitrary, e.g., runtime, solution quality, accuracy or loss.}
	\item The \textit{instance feature} file contains the feature vectors for all instances. Another file contains technical information
	about errors encountered or instances solved during feature computation.
	\item The \textit{cross-validation} file describes how to split the instance set into training and test sets
		to apply a standard machine learning approach to obtain an unbiased estimate of the performance of an algorithm selector.
    \item A human-readable \textit{README} file explains the origin and meaning of the scenario, as well as the process of data generation.
\end{itemize}
\paragraph{Optional Data}
\begin{itemize}
  	\item The \textit{feature costs} file contains the costs of the feature groups\changed{, i.e., sets of features computed together.}
  	\item  The \textit{ground truth} file specifies information on the instances and their 
		respective solutions \changed{(e.g., SAT or UNSAT)}.
	\item The \textit{literature references} file \changed{in BibTeX format} includes information on the context in which the data set was generated 
		and previous studies in which it was used.
\end{itemize}
\end{minipage}
}

\section{Algorithm Selection Scenarios Provided in ASlib Release 2.0}
\label{sec:scenarios}




The set of algorithm selection scenarios in release version 2.0 of our library, shown in Table~\ref{tab:overview},
has been assembled \fh{to represent a diverse set of selection problem settings that covers a wide range of problem domains, types of algorithms, features and problem instances. Our scenarios include both problems that have been broadly studied in the context of algorithm selection techniques (such as SAT and CSP),
as well as more recent ones (such as the container pre-marshalling problem).} 
Most of our scenarios were taken from publications that
report performance improvements through algorithm selection
and consist
of algorithms where the virtual best solver (VBS)\footnote{The VBS is defined as
a solver that perfectly selects the best solver from $\mathcal{A}$ on a per-instance basis.} is 
significantly better than the single best solver.\footnote{The single best
solver has the best performance averaged across all instances.} 
Therefore, these are problems on which it makes sense to seek performance improvements via algorithm selection.
All scenarios are available on our online platform
(\url{http://www.aslib.net/}).

\fh{We now briefly describe the scenarios we included and what makes them interesting.}

\begin{table}
\begin{center}
\small
\begin{tabular}{l cccccc}
\toprule
scenario 				& $\#\mathcal{I}$ 		& $\#\mathcal{A}$ 		& $\#\mathcal{F}$  		& $\#\mathcal{F}_g$  & Costs 	 & Literature \\
\midrule
 SAT$11$-HAND 	& $296$ 	 	& $15$ 	 		& $115$  		& $10$				  & $\checkmark$  & \cite{xu_satzilla_2008}\\
 SAT$11$-INDU 	& $300$			& $18$ 	 		& $115$  		& $10$				  & $\checkmark$ & \cite{xu_satzilla_2008}\\
 SAT$11$-RAND 	& $600$ 		& $9$ 	 		& $115$  		& $10$				  & $\checkmark$ & \cite{xu_satzilla_2008}\\
 SAT$12$-ALL	 	& $1614$		& $31$ 	 		& $115$  		& $10$				  & $\checkmark$ & \cite{xu2012satzilla2012}\\
 SAT$12$-HAND 	& $767$			& $31$ 	 		& $115$  		& $10$				  & $\checkmark$ & \cite{xu2012satzilla2012}\\
 SAT$12$-INDU  	& $1167$		& $31$ 	 		& $115$  		& $10$				  & $\checkmark$ & \cite{xu2012satzilla2012}\\
 SAT$12$-RAND  	& $1362$		& $31$ 	 		& $115$ 		& $10$				  & $\checkmark$ & \cite{xu2012satzilla2012}\\
 SAT$15$-INDU  	& $300$		& $28$ 	 		& $54$ 		& $1$				  & $\times$ & --\\
 QBF-$2011$  	 	&  $1368$		& $5$ 	 		& $46$  		& $1$				  & $\times$ & \cite{pulina_self-adaptive_2009}\\
 QBF-$2014$  	 	&  $1254$		& $14$ 	 		& $46$  		& $1$				  & $\times$ & --\\
 MAXSAT$12$-PMS 	& $876$			& $6$ 	 		& $37$  		& $1$				  & $\checkmark$  & \cite{malitsky_evolving_2013} \\
 MAXSAT$15$-PMS-INDU & $601$			& $29$ 	 		& $37$  		& $1$				  & $\times$  & -- \\
 CSP-$2010$  	 	& $2024$		& $2$ 	 		& $17$  		& $1$				  & $\times$ & \cite{gent_learning_2010}  \\
 CSP-MZN-$2013$		& $4642$		& $11$			& $155$			& $2$				  & $\checkmark$ & \cite{AmadiniGM14} \\
 PROTEUS-$2014$		& $4021$		& $22$			& $198$			& $4$				  & $\checkmark$ & \cite{hurley_proteus_2014}\\
 ASP-POTASSCO 	 	& $1294$		& $11$ 	 		& $138$ 		& $5$				  & $\checkmark$ & \cite{holisc14a}\\
 PREMAR-ASTAR-$2015$ & $527$ & $4$ 	 		& $22$  		& $3$				  & $\times$ & \cite{Ti14tr} \\
\bottomrule
\end{tabular}
\caption{Overview of algorithm selection scenarios in the ASLib with the number of instances $\#\mathcal{I}$, 
number of algorithms $\#\mathcal{A}$, 
number of features $\#\mathcal{F}$,
number of feature processing groups $\#\mathcal{F}_g$
and availability of feature costs.}
\label{tab:overview}
\end{center}
\end{table}

\subsection{SAT: Propositional Satisfiability}


The propositional satisfiability problem (SAT) is a classic NP-complete problem that consists of determining the existence of an assignment of values to variables of a Boolean formula such that the formula is true. It is widely studied, with many applications including formal verification \cite{prasad2005survey}, scheduling \cite{crawford1994experiment}, planning \cite{kautz1999unifying} and graph coloring \cite{van2008another}. Our SAT data mainly stems from different iterations of the SAT competition,\footnote{\url{http://www.satcompetition.org/}} \changed{which is split into three tracks: industrial (INDU), crafted (HAND), and random (RAND).}

\changed{The SAT scenarios are characterized by a high level of maturity and diversity in terms of their solvers, features and instances. Each SAT scenario involves a highly diverse set of solvers, many of which have been developed for several years.} \fh{In addition, the set of SAT features is probably the best-studied feature set among our scenarios; it includes both static and probing features that are organized into as many as ten different feature groups. The instance sets used in our various SAT scenarios range from randomly-generated ones to real-world instances submitted by industry.}

\subsection{QBF: Quantified Boolean Formula}

A quantified Boolean formula (QBF) is a formula in propositional logic with
\changed{universal or existential} quantifiers \changed{on each variable in the} formula. 
A QBF solver finds a set of variable assignments
that makes the formula true or proves that no such set can exist. This is a
\hh{PSPACE-complete} problem\changed{ for which 
solvers exhibit a wide range of} performance characteristics.
Our QBF-2011 data set comes from the QBF Solver Evaluation
2010\footnote{\url{http://www.qbflib.org/index_eval.php}} and consists of instances from the main, small hard, 2QBF and random tracks.
Our QBF-2014 data set comes from the application track of the QBF Gallery 2014\footnote{\url{http://qbf.satisfiability.org/gallery/}}.
The instance features were computed using the AQME system and are described in
more detail by Pulina et al.~\cite{pulina_self-adaptive_2009}.
The solvers for QBF-2011 come from the AQME system as well, whereas the solvers
for QBF-2014 are the ones submitted to the application track of the QBF Gallery.

\subsection{MAXSAT: Maximum Satisfiability}

MaxSAT is the optimization version of the previously introduced SAT problem, and aims to find a variable assignment that maximizes the number of satisfied clauses. The MaxSAT problem representation can be used to effectively encode a number of real-world problems, such as 
FPGA routing~\cite{XuRuSa03}, and software package installation~\cite{ArBeLyMaRa10}, among others, as it permits reasoning about both optimality and feasibility.
The particular scenarios focus on the partial MaxSAT (PMS) problem~\cite{SATHandbook}.


\changed{
The MAXSAT$12$-PMS scenario is composed of a collection of random, crafted and
industrial instances from the 2012 MaxSAT Evaluation~\cite{MAXSATevaluations}.
The techniques used to solve the various instances in this scenario are very
complementary to each other, leading to a substantial performance gap between
the single best and the virtual best solver. Furthermore, \hh{because there are
only six solvers with very different performance characteristics, algorithm
selection} approaches must be very accurate in their choices, as any mistake
is heavily penalized. }

\changedTwo{
The more recent MAXSAT$15$-PMS-INDU was built on the performance data of the
industrial track on partial MAXSAT problems from the 2015 MAXSAT Evaluation.\footnote{\url{http://www.maxsat.udl.cat/15/results/index.html}}
With $29$ algorithms, it provides a larger set of solvers than MAXSAT$12$-PMS.
However, there are different parameterizations of the same solvers,
e.g., four different variants of \emph{ahms},
such that there are some subsets of strongly correlated solvers. 
The performance gap between the single best and virtual
best solver is larger in MAXSAT$12$-PMS than in MAXSAT$15$-PMS-INDU.  
}

\subsection{CSP: Constraint solving}

\changed{Constraint Satisfaction Problem (CSP; ~\cite{StuckeyFSTF14})} is concerned with finding solutions to constraint
satisfaction problems---a task that is NP-complete.
Learning in the context of constraint solving is a technique by which previously unknown
constraints that are implied by the problem specification are uncovered during
search and subsequently used to speed up the solving process.


\changed{
The scenario CSP-$2010$ contains only two solvers: one that employs lazy
learning~\cite{gent_lazy_2010, gent_learning_2010} and one that does not~\cite{gent_minion_2006}.
The data set is heavily biased towards the non-learning solver, such that the baseline (the
single best solver) is very good already. Improving on this is a challenging
task and harder than in many of the other scenarios. Furthermore, both solvers
share a common core, which results in a scenario that directly evaluates the
efficacy of a specific technique in different contexts.
}

\changedTwo{The more recent scenario CSP-MZN-$2013$ provides a larger set of
instances, algorithms and instance features. Instances and algorithms come from
the MiniZinc challenge $2012$ and the International Constraint Solver
Competitions (ICSC) in $2009$. Specifically, the instances come from the
MiniZinc $1.6$ benchmark suite and the algorithms in the scenario participated
in the MiniZinc Challenge $2012$. Algorithms, instances and instance features
are described in more detail in \cite{AmadiniGM14,AmadiniGM14a}.
}


\changedTwo{Our final CSP scenario PROTEUS comes
from}~\cite{hurley_proteus_2014} and includes an extremely diverse mix
of well-known CSP solvers alongside competition-winning SAT solvers 
that have to solve (converted) XCSP instances\footnote{The XCSP instances are taken from
\url{http://www.cril.univ-artois.fr/~lecoutre/benchmarks.html} as described
in~\cite{hurley_proteus_2014}.}. 
The SAT solvers can accept different conversions of the CSP
problem into SAT (see, e.g.,~\cite{daniel_simple_csp_to_sat,tamura_sugar,tanjo_azucar}), 
which in our format are provided as separate algorithms. 
This scenario is the only one in which
solvers are tested with varying ``views'' of the same problem. The
features of this scenario are also unique in that they include both the SAT and
CSP features for a given instance. This potentially provides additional
information to the selection approach that would normally not be available for
solving CSPs. An algorithm selection system has a very high degree of
flexibility here and may choose to perform only part of the possible
conversions, thereby reducing the set of solvers and features, but also reducing
the overhead of performing the conversions and feature computations. There are
also synergies between feature computation and algorithm runs that can be
exploited, e.g., if the same conversion is used for feature computation and to
run the chosen algorithm then the cost of performing the conversion is only incurred
once. In other cases, where features are computed on one representation and
another one is solved, conversion costs are incurred both during feature
computation and the running of the algorithm.

\subsection{ASP: Answer Set Programming}

Answer Set Programming (ASP, \cite{baral02a,gekakasc12a}) is a form of declarative programming 
with roots in knowledge representation, non-monotonic reasoning and constraint solving.
In contrast to many other constraint solving domains (e.g., the satisfiability problem),
ASP provides a rich yet simple declarative modeling language 
in which problems \changed{up to $\Delta_3^{\rm p}$ (disjunctive optimization problems)} can be expressed.
\changed{ASP has proven to be efficiently applicable to many real-world applications},
e.g., product configuration~\cite{soinie99a},
decision support for NASA shuttle controllers~\cite{nobagewaba01a},
synthesis of multiprocessor systems~\cite{ismabogesc09a} 
and industrial team building~\cite{griilelirisc10a}.


\changed{
In contrast to the other scenarios,
the algorithms in the scenario ASP-POTASSCO were automatically constructed by an adapted version of \system{Hydra}~\cite{xu_hydra_2010},
i.e., the set of algorithms consists of complementary configurations of the solver \system{clasp}~\cite{gekasc12b}.
The instance features were generated by a light-weight version of \system{clasp},
including static and probing features organized into feature groups; they
were previously used in the algorithm selector \emph{claspfolio}~\cite{gebser_portfolio_2011,holisc14a}.}

\subsection{PREMAR-ASTAR-2015: Container pre-marshalling}

\changed{The container pre-marshalling problem (CPMP) is an NP-hard container
stacking problem from the container terminals literature~\cite{StVo08}.
\hh{We constructed an algorithm selection scenario from two recent A* and IDA* approaches for
solving the CPMP presented in~\cite{TiPaVo14tr}, using instances from the
literature}. The scenario is described in detail in~\cite{Ti14tr}.

The pre-marshalling scenario \hh{differs from} other scenarios in that the set
of algorithms is highly homogeneous. All of the algorithms are parameterizations
of a single symmetry breaking heuristic, either using the A* or IDA* search
techniques, which stands in sharp contrast to the diversity of solvers present
in most other scenarios.
The scenario represents a real-world,
time-sensitive problem from the operations research literature, where algorithm
selection techniques can have a large impact.}

\section{Automated Exploratory Data Analysis}
\label{sec:eda}

The online platform for our benchmark repository \changedTwo{offers not only}
the scenario data files themselves. It also
provides many tables and figures that summarize them.
These pages are automatically generated and currently consist (among
others) of the following parts:

\begin{itemize}
\item an overview table that lists, for example, the number of instances, algorithms
and features for all available scenarios, similar to Table~\ref{tab:overview};
\item a summary of the algorithms' performance and run status data;
\item a summary of the feature values, as well as the run status and costs of the feature groups;
\item benchmark results for standard machine learning models for each scenario; see Section~\ref{sec:experiments}.
\end{itemize}

Presenting this additional data offers the following advantages:

\begin{itemize}
\item Researchers can quickly understand which scenarios are available and select those best suited to their needs.
\item \replaced{Data can quickly be sanity-checked}{Data can be sanity-checked by eye-balling}. It is common that data collection errors occur when scenario data is gathered and submitted for the first time.
\item Interesting or challenging properties of the data sets become visible, providing the researcher with a quick and informative first impression.
\end{itemize}

The \replaced{platform's}{} \textbf{summary page for the algorithms} starts with
a table listing summary statistics
regarding their performance (e.g., mean values and standard variations) and run status 
(e.g., how many runs were successful). 
We also indicate whether one algorithm is dominated by another, i.e., an algorithm $a_1$ dominates another algorithm $a_2$ if and only if $a_1$ has performance at least equal to that of $a_2$ on all instances, and $a_1$ outperforms $a_2$ on at least one instance.
This is useful, because there is no reason to include a dominated algorithm in a portfolio.
Various visualizations, such as box plots, scatter plot matrices, correlation plots and density plots 
enable further inspection of the performance distribution
and correlation between algorithms\changed{, allowing the reader to better understand the strengths and weaknesses of each algorithm.} \replaced{}{For display, we impute high values for the missing or \changedTwo{censored} performance values corresponding to failed runs so that they are clearly visible rather than silently discarded.} All of our plots can be configured to use log scales, which often improves visual understanding \changedTwo{of heavy-tailed distributions (e.g., runtime distributions of hard combinatorial solvers~\cite{GomesSCK00})}.

\begin{figure}[!t]
	\begin{minipage}[!t]{0.29\textwidth}
	\includegraphics[width=\textwidth]{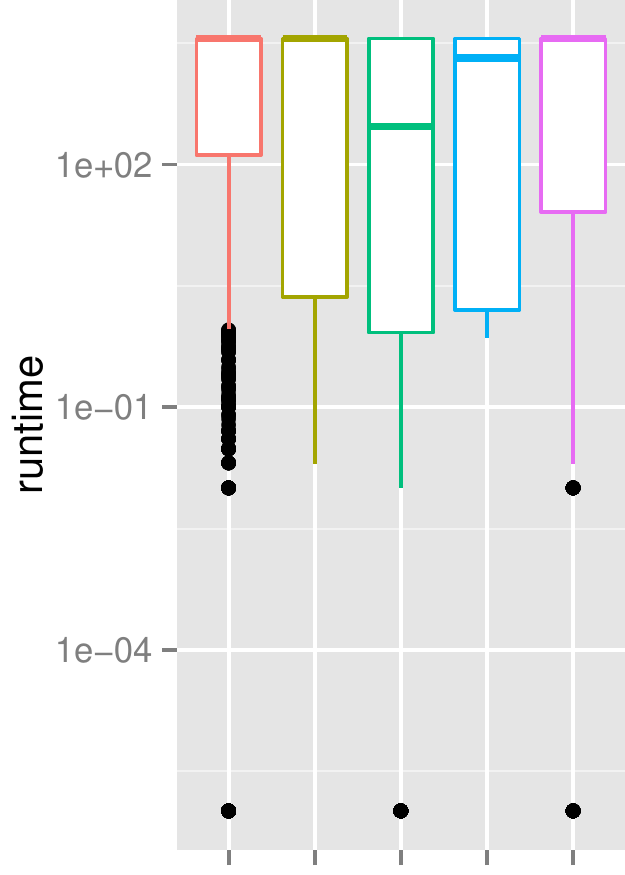}
	\end{minipage}
	\hfill
	\begin{minipage}[!t]{0.7\textwidth}
	\includegraphics[width=\textwidth]{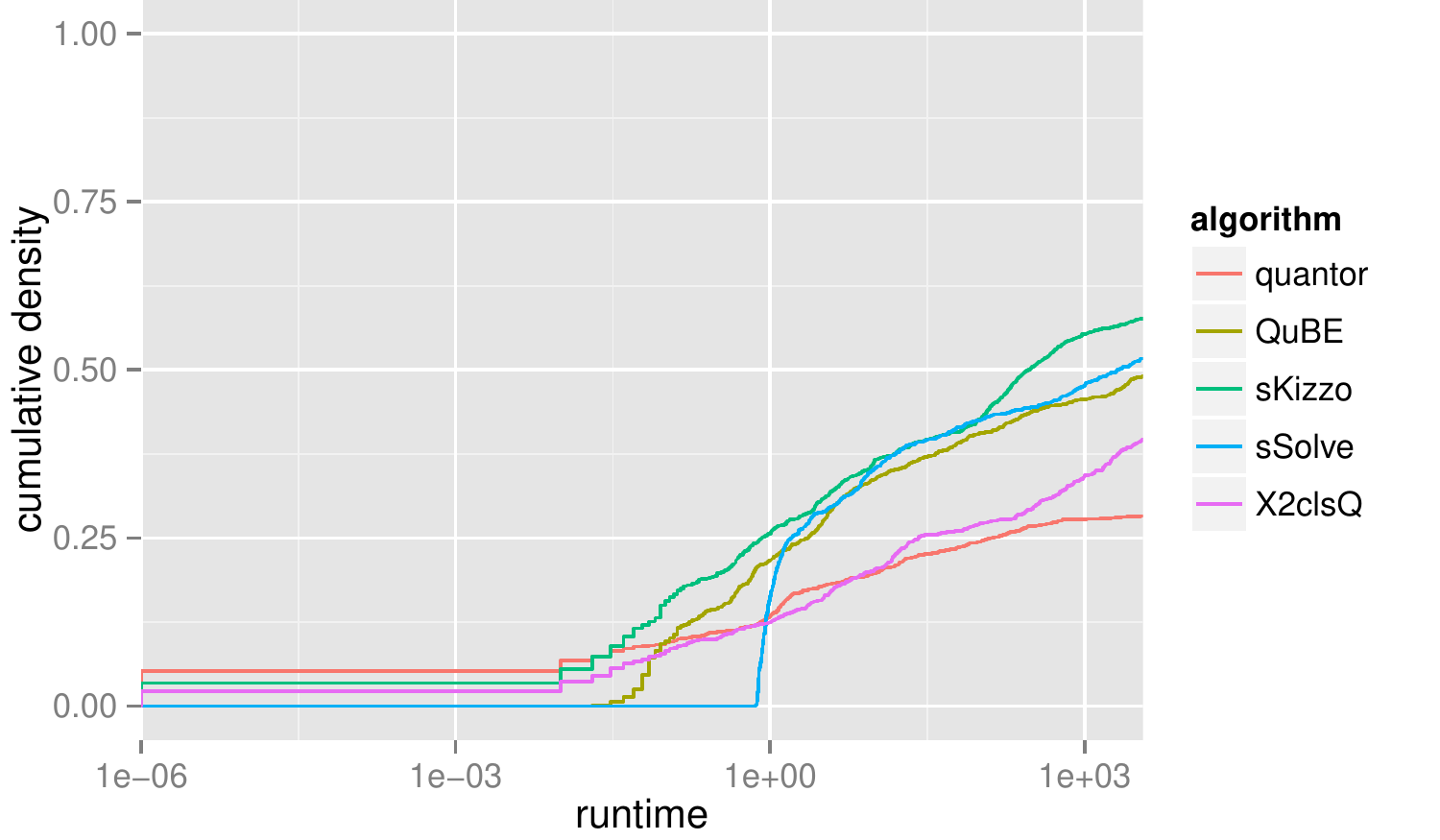}
	\end{minipage}
	 \caption{\changed{Algorithm performance distributions of the QBF-$2011$
     scenario: Boxplots (left) and cumulative distribution functions (right);
     both on a log scale.}}
	\label{fig:qbf}
\end{figure}


Figure~\ref{fig:qbf} shows \changed{boxplots} and cumulative distribution
functions for the algorithms in the QBF-$2011$ scenario as an example.
\replaced{The boxplots
summarize the runtimes of an algorithm by drawing a box between the 25\%- and
75\%-quantile of the sample, i.e.,~the smallest values that are greater or equal
to 25\% and 75\% of the runtimes. In addition, each box contains a line showing
the median runtime, as well as so-called whiskers, i.e.,~lines that connect the
box with runtimes that are within 150\% of the interquartile range (the length
of the box) below the 25\%- or above the 75\%-quantile, respectively.
Observations with even more extreme runtimes are considered to be outliers and
are depicted by a single point per outlier. The
cumulative distribution function plots on the other hand
show runtimes on all instances for the algorithm. Each point within the plot consists of
the observed runtime on the x-axis and the corresponding cumulative density,
i.e., the percentage of instances that were solved at this or a smaller runtime,
on the y-axis.}{}

Such plots show the location of the mean, distribution spread, density
multimodality and other properties of the distribution. In addition, they reveal
how long it took an algorithm to solve the given instances. \changed{For
example, for the QBF-2011 scenario in Figure~\ref{fig:qbf}, one can see that the
algorithm \system{quantor} finds a solution very quickly on a few instances,
i.e., it solves approximately $5\%$ of the instances \changedTwo{in less than a
second.} However, if it does not succeed quickly, it often does not succeed at
all---it solved less than $30\%$ of all the instances. In contrast,
\system{sSolve} usually needs longer to find a solution, but by the time it
does, it is one of the best algorithms. Such behavior can indicate that the
algorithm \hh{requires a `warm-up'} stage, which should be considered when
deploying it.}

\begin{figure}[!t]
	\centering
	\begin{minipage}[!t]{0.55\textwidth}
	\includegraphics[width=1.05\textwidth]{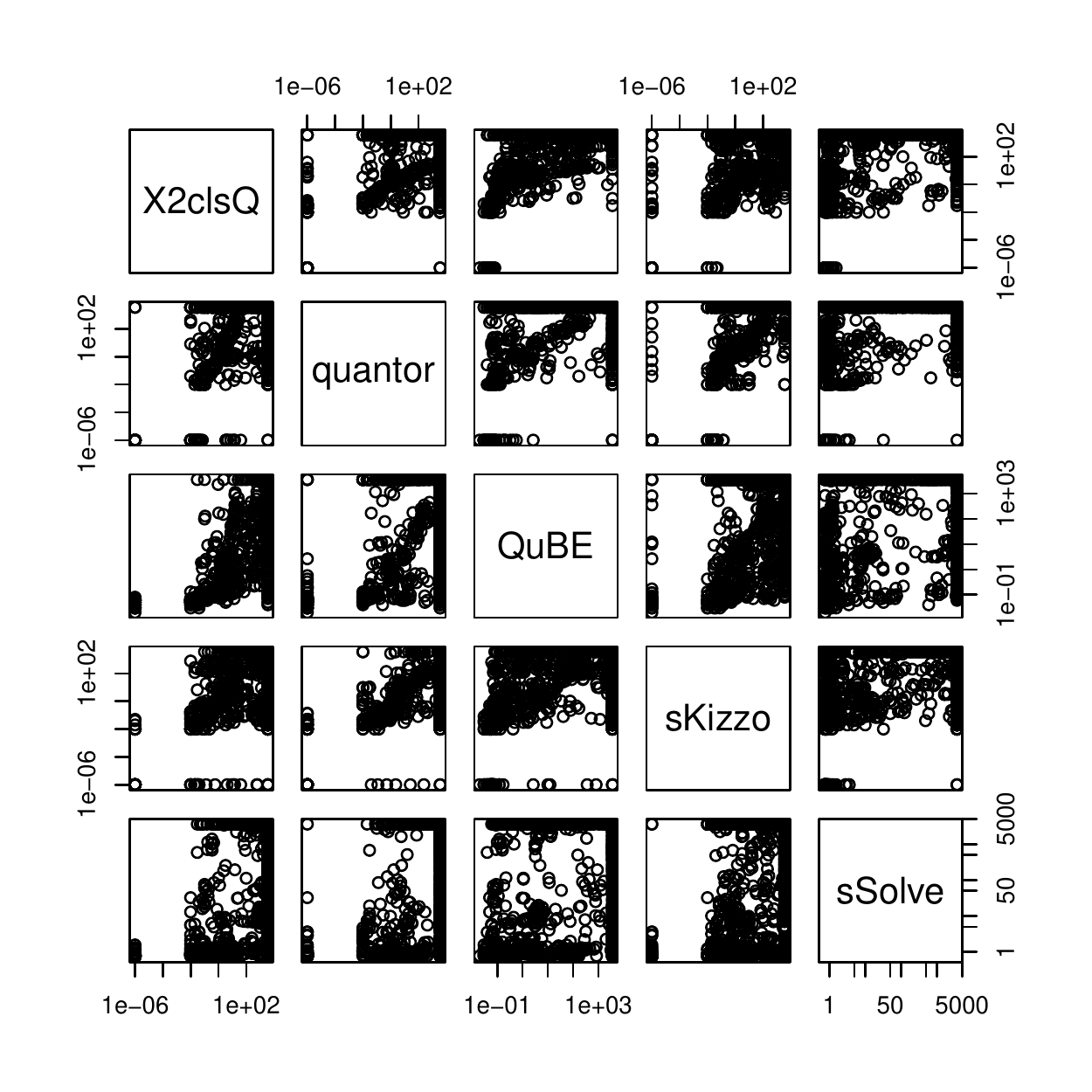}
	\end{minipage}
	\hfill
	\begin{minipage}[!t]{0.43\textwidth}
	\hspace{-0.65cm}
	\includegraphics[width=1.15\textwidth]{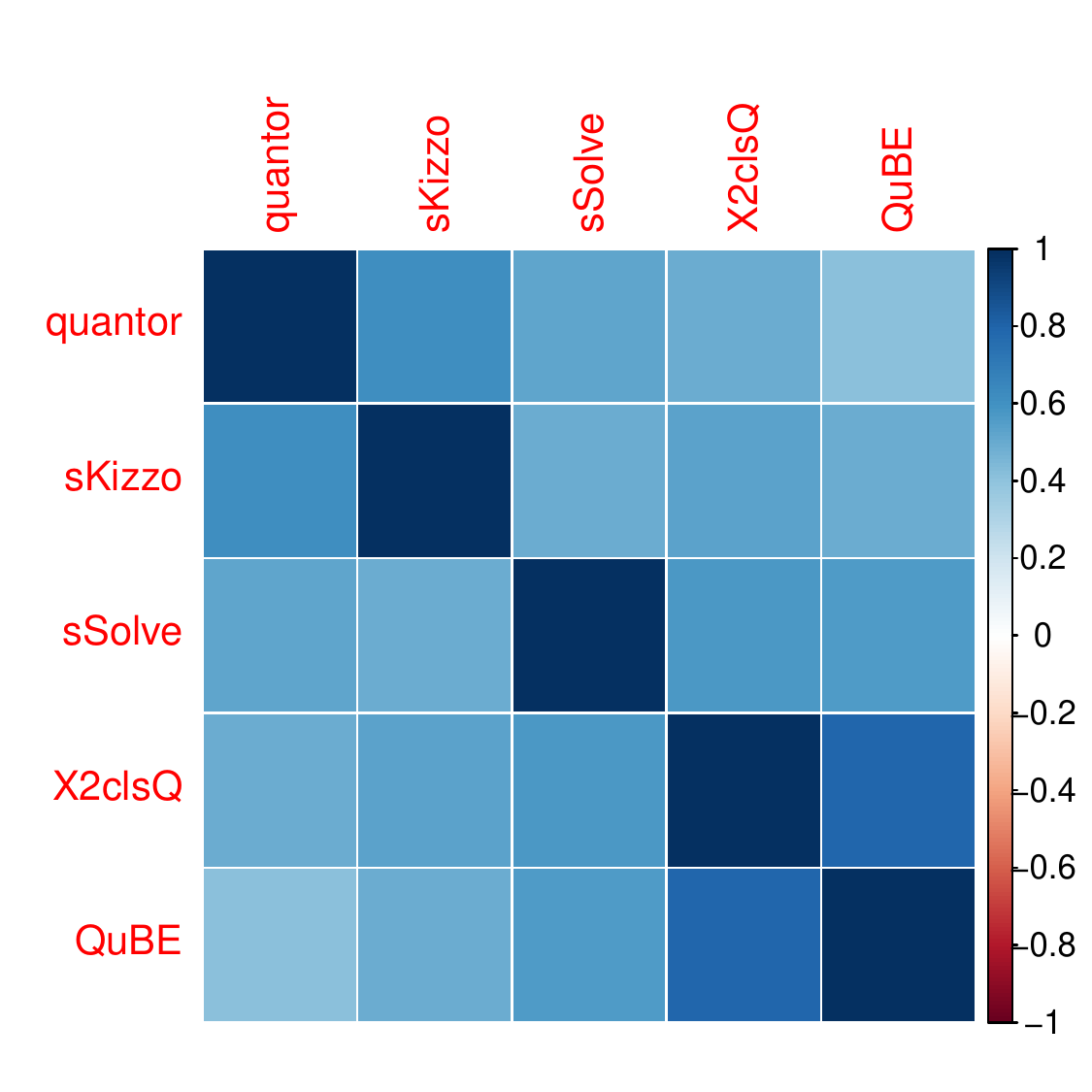}
	\end{minipage}
	 \caption{\changed{Pairwise correlations among algorithms of the QBF-$2011$ scenario: A scatter plot matrix on a log scale (left) and the plot of a correlation matrix (right).}}
	\label{fig:scatter}
\end{figure}

The left panel of \changed{Figure~\ref{fig:scatter} shows pairwise scatterplots of the QBF-$2011$ scenario, allowing an easy comparison of algorithm pairs on all instances from a given scenario. 
Each point represents a problem instance within the scenario, and from the location
of the point cloud one can see whether an algorithm is dominant over the
majority of instances, or whether relative performance strongly varies across instances. The first case can be identified by a cloud that is located either in the upper-left or lower-right corner of a single scatterplot. In such a case, the dominated algorithm could be discarded from the portfolio. However, if this type of \replaced{dominance relationship}{domination} is not present, there is the potential to realize performance improvements by means of per-instance algorithm selection.}


\changed{\hh{Because detecting correlation in algorithm performance is also of
interest when analyzing the strengths and weaknesses of a given portfolio-based
solver}~\cite{xu2012evaluating},} we also present a correlation
matrix, cf. Figure \ref{fig:scatter} (right panel). \changed{Algorithms that
have a (high) positive correlation are more likely to be redundant in a
portfolio, whereas pairs with a (high) negative correlation are more likely to
complement each other.} We calculate Spearman's correlation coefficient between
ranks. Blue boxes represent positive correlation, red boxes represent negative
correlation, and shading indicates the strength of correlation. The algorithms
are also clustered according to these values (using Ward's method~\cite{ward63})
and then sorted, such that similar algorithms appear together in blocks.
\changed{This type of clustering allows the identification of algorithms with
highly correlated performance.}

\begin{figure}[!t]
	\includegraphics[width=\textwidth]{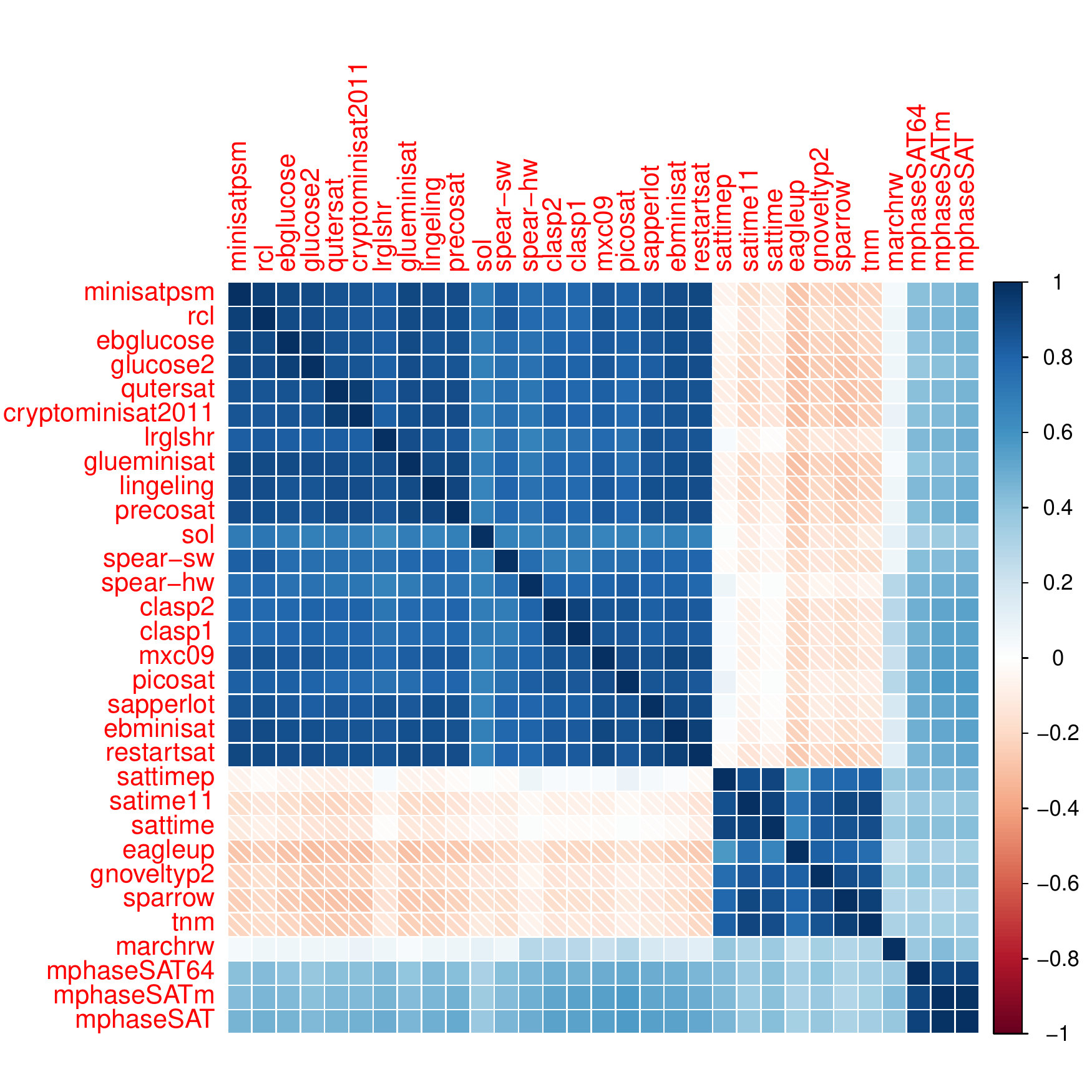}
	 \caption{Algorithm correlations on the SAT$12$-ALL scenario.}
	\label{fig:cormat}
\end{figure}

\changed{\replaced{Figure~\ref{fig:cormat} shows the correlation between
algorithms for the SAT$12$-ALL scenario. The plot reveals}{The example given in
Figure \ref{fig:cormat} (the plot of a correlation matrix of the SAT$12$-ALL
scenario) shows} four groups of algorithms (\system{minisatpsm} to
\system{restartsat}, \system{sattimep} to \system{tnm}, \system{marchrw} and the
three \system{mphaseSAT}-algorithms) with high correlations within each group.
It may be desirable to include just a single representative from each group,
reducing the size of the entire portfolio from 31 to four algorithms.}

As we do with algorithm runs, we characterise the features by giving summary statistics of the
feature values, the run status and \changedThree{the} cost of the feature groups.
Table~\ref{table:features} shows the summary of the feature groups for the
SAT$12$-RAND scenario. \changed{In this scenario, all 115 features have the
feature group ``Pre'' as a requirement.
While this preprocessing group
succeeded in all cases, \fh{one other group did not: the feature group ``CG'' (which computes clause graph features) failed in 37.37\% of cases due to exceeding time or memory limits, and even for instances where it succeeded, it was quite expensive ($8.79$ seconds on average).} 
Such information is useful to understand the behavior of the features:
\changedThree{how risky is it to compute a feature group, and how much time must one invest in order to obtain the corresponding features?}}

\begin{table}[t]
\centering
{\small
\begin{tabular}{rrrrrrrrr}
\toprule
& & \multicolumn{3}{c}{runstatus [\%]} & \multicolumn{4}{c}{cost [s]} \\ 
\cmidrule(lr){3-5}\cmidrule(lr){6-9}
feature group & $\#$ & ok & $\ldots$ & crash & min & mean & max & missing [\%] \\ 
  \midrule
  Pre 		& 115 & 100.00	& $\ldots$ & 0.00 & 0.00 & 0.06 & 1.31 & 0.00 \\ 
  Basic 		&  14 & 100.00 	& $\ldots$ & 0.00 & 0.00 & 0.00 & 0.07 & 0.00 \\ 
  KLB 		&  20 & 100.00 	& $\ldots$ & 0.00 & 0.00 & 0.18 & 6.09 & 0.00 \\ 
  CG 		&  10 & 62.63 	& $\ldots$ & 37.37 & 0.02 & 8.79 & 20.28 & 0.00 \\ 
  DIAMETER	&   5 & 100.00 	& $\ldots$ & 0.00 & 0.00 & 0.60 & 2.11 & 0.00 \\ 
  cl 		&  18 & 100.00 	& $\ldots$ & 0.00 & 0.01 & 1.99 & 2.02 & 0.00 \\ 
  sp 		&  18 & 100.00 	& $\ldots$ & 0.00 & 0.01 & 0.33 & 3.05 & 0.00 \\ 
  ls\_saps 	&  11 & 100.00 	& $\ldots$ & 0.00 & 1.36 & 2.12 & 2.51 & 0.00 \\ 
  ls\_gsat 	&  11 & 100.00 	& $\ldots$ & 0.00 & 2.03 & 2.29 & 3.03 & 0.00 \\ 
  lobjois 	&   2 & 100.00 	& $\ldots$ & 0.00 & 2.00 & 2.00 & 2.27 & 0.00 \\ 
   \bottomrule
\end{tabular}
}
\caption{Feature group summary for the SAT$12$-RAND scenario. The second column shows how many features
  depend on another feature group to be computed first. Percentages of runstatus
  events are followed by summary statistics for group costs.}
\label{table:features}
\end{table}


We also check whether instances occur with exactly the same feature values, indicating that the experimenter
might have erroneously run on the same instance twice. 

All of the above tables and figures and many more were generated by our online
platform, and are also accessible through the R package \system{aslib}. The
functions are highly configurable and customisable.
\changed{We plan to extend our data analysis with additional techniques, such as
more measures of algorithm performance~\cite{SmithMiles201412}.}


\section{\changedTwo{Study of Algorithm Selection Techniques}}
\label{sec:experiments}


\fh{In this section, we present an exploratory benchmark study that gives an indication of the diversity of our benchmarks. First, we evaluate the performance
of algorithm selectors on our scenarios. We then perform a subset selection study to identify the important algorithms and instance features in each of the scenarios.
We make no claim that the presented experimental settings are exhaustive or that we achieve state-of-the-art 
algorithm selection performance; rather, we provide results that can be used as
a baseline comparison for other approaches.
These results, and our
framework in general, allow us to study which algorithm selection approaches work
well for which of our scenarios.}


We use the \system{LLAMA} toolkit~\cite{kotthoff_llama_2013}, version 0.9.1, in combination with the \system{aslib}
package\footnote{\url{https://github.com/coseal/aslib-r}} to
run the algorithm selection study. \system{LLAMA} is an R~\cite{R}
package
that facilitates many common algorithm selection approaches. In particular, it
enables access to classification, regression, and clustering models for algorithm
selection---the three main approaches we use in our study.
We use the \system{mlr} R
package~\cite{bischl_mlr} as an interface to the machine learning models
provided by other R packages. We parallelize all of our benchmark experiments
through the \system{BatchExperiments} \cite{Bischl2015_1} R package.

In this paper, we only present aggregated benchmark results, but the
interested reader can
access full benchmark results at \url{http://aslib.net}. Our
study is fully reproducible as the complete code to generate these
results is part of the \system{aslib} package.

We use the subset of feature groups that are recommended by the authors of the
respective scenario, called default feature set. For the feature subset selection study, we have used all
feature groups. Detailed and continuously updated information (e.g., the names
of the feature processing groups we selected and their average costs) is
provided on the ASlib website.

\subsection{Experimental setup}

We consider three approaches to algorithm selection that have been studied extensively in the
literature (cf.\ Section~\ref{sec:background:how}): 

\begin{itemize}
  \item \emph{classification} applies a multi-class classifier to directly predict the \changedTwo{best performing algorithm} of the $k$ possible algorithms;
  \item \emph{regression} predicts each algorithm's performance via a regression model and then chooses the one with the best predicted performance;
  \item \emph{clustering} groups problem instances in feature space, then
determines the cost-optimal solver for each cluster and finally assigns to each
new instance the solver associated with the instance's predicted cluster.
\end{itemize} 

\begin{table}[thb]
\centering
\small
\begin{tabular}{p{0.2em}llr}
\toprule[1pt]
& Technical Name & Algorithm and Parameter Ranges & Reference\\
\midrule
\multicolumn{2}{l}{\emph{classification}}\\
& ksvm & support vector machine & \cite{ksvm}\\
& & $C\in [2^{-12}, \; 2^{12}], \; \gamma \in [2^{-12}, \; 2^{12}]$\\
& randomForest & random forest  & \cite{rf}\\
& & $\texttt{ntree}\in [10, \; 200], \; \texttt{mtry}\in [1, \; 30]$\\
& rpart & recursive partitioning tree, CART & \cite{rpart}\\
\midrule
\multicolumn{2}{l}{\emph{regression}}\\
& lm & linear regression & \cite{R}\\
& randomForest & random forest & \cite{rf}\\
& & $\texttt{ntree} \in [10, \; 200], \; \texttt{mtry} \in [1, \; 30]$\\
& rpart & recursive partitioning tree, CART & \cite{rpart}\\
\midrule
\multicolumn{2}{l}{\emph{clustering}}\\
& XMeans & extended $k$-means clustering & \cite{hall_weka_2009}\\
\bottomrule[1pt]
\end{tabular}
\caption{Machine learning algorithms \pk{and their parameter ranges} used for our study.}
\label{tab:mlalgs}
\end{table}



The specific machine learning algorithms we employed for our study are
shown in Table~\ref{tab:mlalgs}. They include
representatives of each of the three major approaches above. 

\changedTwo{The linear model we employ is the best-studied regression method. In its most basic version, it models the data using the linear function $f(x) = \beta^T x + \beta_0$; parameters are obtained by minimizing squared loss. 
The trees constructed by the CART algorithm, which can handle both classification and
regression problem, are grown in a top-down manner and divide the training
data into rectangular regions by axis-parallel splits at each interior node.
Splits are selected by considering label impurity reduction measured by an
impurity function, based, e.g., on the Gini index for classification or squared
loss for regression. Leaf nodes associate the best, but constant, label with
their feature region for prediction. 
Random forests form an ensemble of $ntree$ simpler trees by bootstrapping
multiple data sets from the original one and then fitting a tree for each.
Predictions are made through majority voting for classification and averaging for
regression. Furthermore, ensemble members are decorrelated by randomly selecting
only a few candidate features for each split point (controlled by parameter
$mtry$) in a tree and maximally growing trees without any early stopping or
pruning.
Support Vector Machines perform linear classification in a transformed feature space by maximizing the margin between the positive and negative examples. Parameter $C$ controls the trade-off between the size of the margin and classification loss.
The feature mapping is implicitly built into the algorithm by substituting the regular inner product of the Euclidean space with a so-called kernel. Parameter $\gamma$ is a property of the radial basis function kernel used here. 
The XMeans clustering algorithm is the only unsupervised learning algorithm we
study. It is an extension of the well known $k$-means method to adaptively
select the number of clusters. $k$-means starts with $k$ random cluster centroids,
assigns each point to the nearest centroid, and then iteratively recomputes the
cluster centroids and cluster assignments until convergence. For further details
on all methods the reader is referred to the standard literature~\cite{Hastie2009} and for XMeans to~\cite{Pelleg2000}.
}

We tuned the hyperparameters of ksvm and randomForest (classification and
regression) within the listed parameter ranges, using random search with 250
iterations and a nested cross validation (with three internal folds) to
ensure unbiased performance results.
All other parameters were left at their default values.
For the clustering algorithm, we set the (maximum) number of clusters to 30
after some preliminary experiments; the exact number of clusters is determined
internally by XMeans.

\subsection{Data preprocessing}

We preprocessed the data as follows.
We removed constant-valued (and therefore irrelevant) features and imputed
missing feature values as the mean over all non-missing values of the
feature.\footnote{For the CSP-MZN-2013 scenario, we also removed the
\texttt{gc\_circuit} feature, which is almost constant.}
\changedTwo{For the clustering methods, we normalized the range of each feature
to the interval $[-1,1]$. The scenarios we consider in this article contain only
continuous features.} The other machine learning methods that require normalized
data perform this internally (for example the SVMs).
\pk{Missing performance values were imputed using the timeout value of the
scenario.}

For each problem instance, we calculated the total feature computation cost
as the sum of the costs of all feature groups, in the order specified in the
definition of the scenario.
If the problem
instance was solved during feature computation \changedTwo{(e.g., using
SLS-probing features~\cite{xu_satzilla_2008})}, we only considered the cost of
the feature groups up
to the one that solved it. Furthermore, we set the runtime
for all algorithms to zero for instances solved during feature
computation. If the instance was not solved during feature computation, we added the feature costs computed in this way to the runtimes
of the individual algorithms on the respective instances \changedTwo{($c(f_{i}) + t(i,a)$)}. 
Given these new runtimes, we checked whether the specified timeout was now
exceeded and set the run status of any corresponding algorithm accordingly. 
Preprocessing runtimes to include feature computation time in this way allows us
to focus on an algorithm selection system's overall performance, and avoids overstating the fraction of instances that would be solved within a time budget in cases where features are expensive to compute. 

Each scenario specifies a partition into $10$ folds for cross-validation to
ensure consistent evaluation
across different methods. We used this partition in our study.

\subsection{Evaluation}

\changedTwo{\noindent We evaluated different algorithm selection models using three different
measures:} 

\begin{itemize}
  \item the fraction of all instances solved within the timeout;
  \item the penalized average runtime with a penalty factor of 10, i.e., a
  timeout counts as 10 times the timeout;
  \item the average misclassification penalty, which, for a given instance, is the difference between the performance of the selected algorithm and the performance of the best algorithm. 
\end{itemize}

\pk{The} performance of each algorithm selection model \pk{was compared} to the virtual best
solver (VBS) and the single best solver. The virtual best solver selects the
best solver from $\mathcal{A}$ for each instance
\changedTwo{($\forall i \in \mathcal{I}: \argmax_{a \in \mathcal{A}} m(i,a)$)}.
Note that the misclassification penalty for VBS is
zero by definition. The single best solver is the actual solver that has the
overall best performance on the data set
\changedTwo{($\argmax_{a \in \mathcal{A}} \sum_{i \in \mathcal{I}} m(i,a)$)}.
Specifically, we consider the solver
with the best PAR10 score over all problem instances in a scenario.%

\subsection{Results}

\begin{figure}[t]
\includegraphics[width=\textwidth]{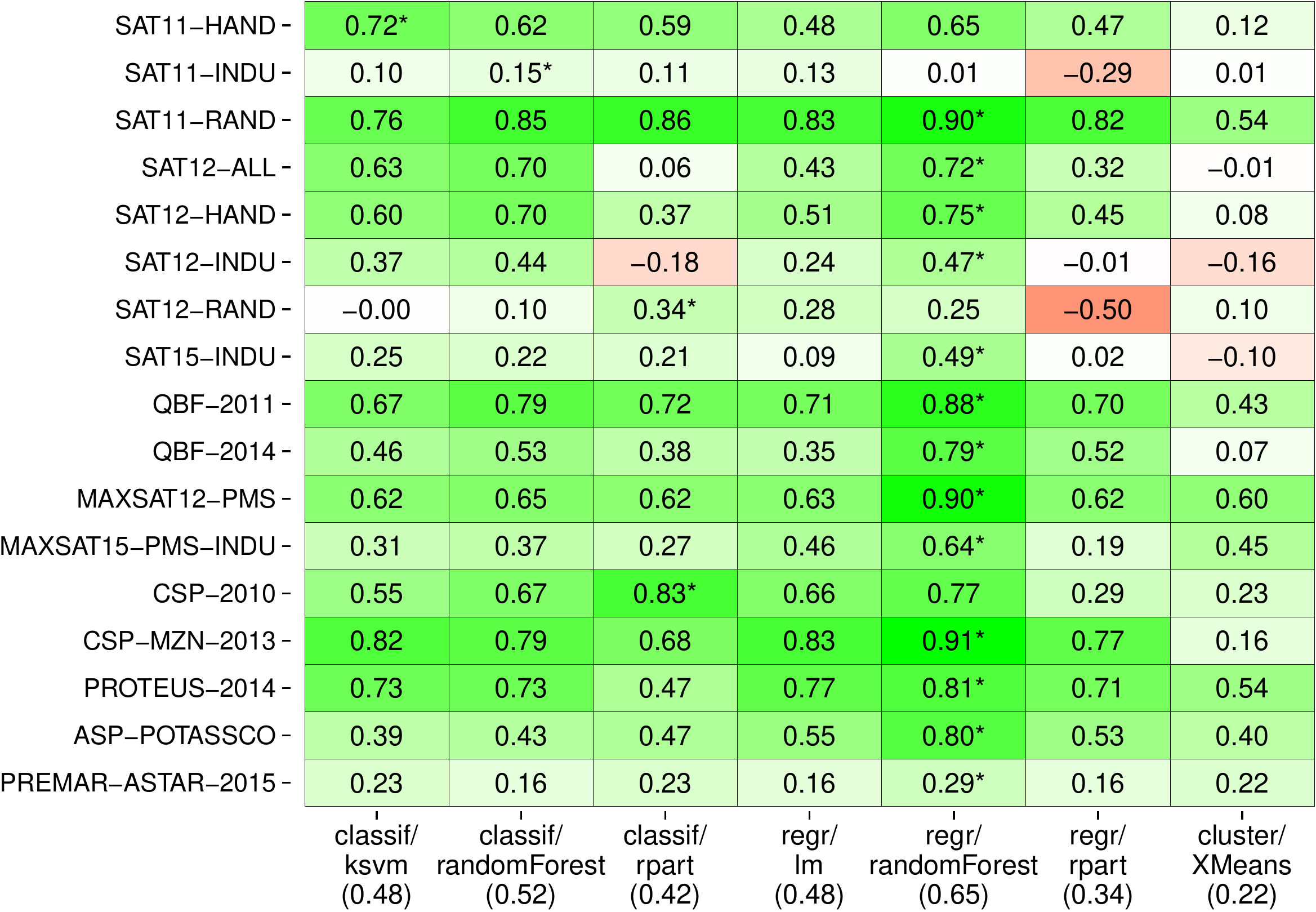}
\caption{Summary of the results of our study. 
\pk{We show how much of the gap between the single best and the virtual best
solver in terms of PAR10 score was closed by each model. That is, a value of 0
corresponds to the single best solver and a value of 1 to the virtual best.
Negative values indicate performance worse than the
single best solver. Within each data set, the best model is annotated with an
asterisk.
The shading emphasizes that comparison: green
cells correspond to values close to 1 (i.e.,\ close to the virtual best solver),
whereas red shows the models with bad performance. White shading indicates
values close to 0, i.e.\ the model has the same performance as the single best
algorithm. The arithmetic mean for each model type
across all scenarios is given in parentheses after the model name.}
}
\label{fig:res-heat}
\end{figure}

Figure~\ref{fig:res-heat} presents a summary of the results of our study. In
most cases, the algorithm selection approaches performed better than the single
best solver. We expected this, as most of our data sets come from 
publications that advocate algorithm selection systems.

Nevertheless, there were significant differences between the scenarios. 
While almost all algorithm selection approaches outperformed the single
best algorithm, there are some scenarios that seem to be much harder for algorithm
selection. In particular, on the \pk{SAT12-INDU} scenario, \pk{three}
approaches were not able to achieve a performance improvement.

Random regression forests stood out \changedThree{quite clearly as} the best overall approach,
achieving the best performance on $13$ of the $17$ datasets. This is in line
with recent results showing the strong performance of this model for algorithm
runtime prediction~\cite{hutter2014algorithm}. The results are also consistent
with those of the original papers introducing the datasets.

XMeans performed worst on average.
On some scenarios, it performed well, in particular SAT11-RAND,
MAXSAT12-PMS, and PROTEUS-2014.
However, on SAT12-ALL, SAT12-INDU, and SAT15-INDU
XMeans performed worse than the single best solver. 
The default subset of instance features appears to be unfavorable for XMeans on industrial SAT instances.



\subsection{Algorithm and Feature Subset Selection}

To provide further insight into our algorithm selection scenarios,
we applied forward selection~\cite{Kohavi97} to the algorithms and features to
determine whether smaller subsets still achieve comparable performance.
We performed forward search independently for algorithms and features for each
scenario. \changedTwo{Forward selection is an iterative selection algorithm whose the first iteration starts with an empty set of algorithms and features; in each subsequent iteration, it greedily adds the 
algorithm or feature to the set which most improves the cross-validated score (PAR$10$) of the predictor.}
The selection process terminates when the score does not improve by at least
$\varepsilon$. We set $\varepsilon = 1$, which roughly corresponds to an
improvement of 1 second per instance.
In all other aspects, the experimental setup was the same as described before.

We used random regression forests\footnote{We
used random forests with default parameters, as the tuning was done for the
set of features specified by the scenario authors and the full set of solvers.}, as it was the best overall approach so far.
We note that the selection results use \changedTwo{standard cross validation rather} than the 
nested version, which may result in \fh{overconfident performance estimates for
the selected subsets~\cite{Bischl2012_3}}.
We accept this caveat since our goal here is to study the ranking of the features and the size of the selected sets,
and a more complex, nested approach would have resulted in multiple selected sets.


\begin{table}[thb]
\begin{center}
\begin{tabular}{lccc}
\toprule
Scenario & PAR10 full set & Number & PAR10 reduced set \\ 
  \midrule
SAT11-HAND & 16943.49 & 15 $\to$ 8 & 15919.09 \\ 
  SAT11-INDU & 13246.70 & 18 $\to$ 4 & 12127.05 \\ 
  SAT11-RAND & 10253.09 & 9 $\to$ 4 & 10180.39 \\ 
  SAT12-ALL & 971.45 & 31 $\to$ 11 & 979.17 \\ 
  SAT12-HAND & 4303.81 & 31 $\to$ 10 & 4187.13 \\ 
  SAT12-INDU & 2763.37 & 31 $\to$ 7 & 2775.61 \\ 
  SAT12-RAND & 3241.42 & 31 $\to$ 2 & 3153.48 \\ 
  SAT15-INDU & 3845.52 & 28 $\to$ 6 & 3604.57 \\ 
  QBF-2011 & 9232.49 & 5 $\to$ 4 & 9198.01 \\ 
  QBF-2014 & 2090.02 & 14 $\to$ 8 & 2040.33 \\ 
  MAXSAT12-PMS & 3370.22 & 6 $\to$ 3 & 3299.15 \\ 
  MAXSAT15-PMS-INDU & 1752.57 & 29 $\to$ 7 & 1479.31 \\ 
  CSP-2010 & 6541.20 & 2 $\to$ 2 & 6516.57 \\ 
  CSP-MZN-2013 & 4204.58 & 11 $\to$ 9 & 4168.35 \\ 
  PROTEUS-2014 & 5905.71 & 22 $\to$ 8 & 5725.50 \\ 
  ASP-POTASSCO & 525.55 & 11 $\to$ 5 & 509.61 \\ 
  PREMAR-ASTAR-2015 & 5154.40 & 4 $\to$ 3 & 4954.45 \\ 
   \bottomrule

\end{tabular}
\caption{PAR$10$ scores for the set of all algorithms and the reduced set, along
with the number of all algorithms and the size of the reduced portfolio.}
\label{tab:asel}
\end{center}
\end{table}    

\begin{table}[thb]
\begin{center}
\begin{tabular}{lcccc}
\toprule
Scenario & full set & default set & Number & reduced set \\ 
  \midrule
SAT11-HAND & 17249.59 & 16943.49 & 113 $\to$ 6 & 15743.04 \\ 
  SAT11-INDU & 13111.61 & 13246.70 & 112 $\to$ 4 & 10951.00 \\ 
  SAT11-RAND & 10496.39 & 10253.09 & 112 $\to$ 3 & 9854.11 \\ 
  SAT12-ALL & 994.25 & 971.45 & 113 $\to$ 9 & 815.37 \\ 
  SAT12-HAND & 4298.00 & 4303.81 & 113 $\to$ 6 & 4092.58 \\ 
  SAT12-INDU & 2881.97 & 2763.37 & 113 $\to$ 6 & 2506.25 \\ 
  SAT12-RAND & 3196.28 & 3241.42 & 113 $\to$ 3 & 3088.72 \\ 
  SAT15-INDU & 3970.56 & 3845.52 & 51 $\to$ 3 & 3620.06 \\ 
  QBF-2011 & 9229.99 & 9232.49 & 46 $\to$ 5 & 8995.62 \\ 
  QBF-2014 & 2102.79 & 2090.02 & 46 $\to$ 4 & 2032.50 \\ 
  MAXSAT12-PMS & 3321.22 & 3370.22 & 30 $\to$ 4 & 3296.52 \\ 
  MAXSAT15-PMS-INDU & 1696.69 & 1752.57 & 29 $\to$ 5 & 1520.77 \\ 
  CSP-2010 & 6514.37 & 6541.20 & 69 $\to$ 3 & 6415.23 \\ 
  CSP-MZN-2013 & 4192.82 & 4204.58 & 115 $\to$ 5 & 4119.36 \\ 
  PROTEUS-2014 & 6120.13 & 5905.71 & 193 $\to$ 6 & 5700.05 \\ 
  ASP-POTASSCO & 516.47 & 525.55 & 134 $\to$ 4 & 472.84 \\ 
  PREMAR-ASTAR-2015 & 5289.96 & 5154.40 & 22 $\to$ 3 & 4619.49 \\ 
   \bottomrule

\end{tabular}
\caption{PAR$10$ scores for the set of all features, the default feature set and the reduced set, along
with the number of all features and the size of the reduced feature set.}
\label{tab:fsel}
\end{center}
\end{table}    

Tables~\ref{tab:asel} and~\ref{tab:fsel} present the results of forward selection for
algorithms and features on all scenarios.
Usually, the number of selected features is very small compared to the complete
feature set.
This is consistent with the observations of \changedTwo{\emcite{RobertsH09} and} \emcite{hutter_identifying_2013} 
who found in their experiments that only a few instance features are necessary
to reliably predict the runtime of their algorithms. For example, on SAT12-RAND,
the only three features selected were
a balance feature concerning the ratio of positive and negative occurrences of
each variable in each clause and
two features based on survey propagation.

The number of algorithms after forward selection is also substantially reduced
on most scenarios. On the SAT scenarios, we expected to see this because the
scenarios consider a huge set of SAT solvers that were not pre-selected in any
way. \emcite{xu2012evaluating} showed  
that many SAT solvers are strongly correlated 
and make only very small contributions to the VBS, \changedTwo{a finding that is corroborated by our results (see Figure \ref{fig:cormat} in Section~\ref{sec:eda})}. For
example, on the SAT12-RAND scenario, only two solvers were selected: sparrow and
eagleup.
We did not expect the set of algorithms to be reduced on the
ASP-POTASSCO scenario, as the portfolio was automatically constructed using
algorithm configuration to obtain a set of complementary parameter settings that
are particularly amenable to portfolios; nevertheless only 5 of 11
configurations were chosen by the forward selection.


Our results indicate that in real-world settings, selecting the most predictive
features and the solvers that make the highest contributions can be important.
More detailed and continuously updated results can be found on the ASlib
website.


\section{\changedTwo{Competitions on ASlib}}
\label{sec:comp}

As described before and illustrated in Section~\ref{sec:experiments}, 
we \changedThree{designed ASlib to enable easy and fair comparison of different algorithm selection approaches}.
The next step to get unbiased performance comparisons of algorithm selectors is to organize competitions based on ASlib.
In this section, we will briefly describe two exemplary competition settings based on ASlib.

\paragraph{On-going Evaluation on ASlib}

In the on-going evaluation on ASlib\footnote{%
The most recent results of the on-going evaluation can be found on the ASlib homepage \url{aslib.net}.},
every participant can simply submit his/her performance for each scenario (using the provided cross validation splits) 
and the source code of their algorithm selector.
The latter will \changedThree{only be} used to verify the results in case of doubt.
The results (i.e., (penalized) average performance on each scenario) will be added in an overview table and the system will be linked. 

In this setting, every system that can read the ASlib format can easily participate and no deadlines for submission are required.
Therefore, the newest systems and results can \changedThree{always be} added on-the-fly 
such that the on-going evaluation always reflects the most recent known state-of-the-art systems and their performances.
Disadvantages of this setting are: 
\begin{enumerate}
  \item the different participants use different amounts of computational resources to compute the results -- for example, two well-performing systems in the on-going evaluation are \emph{SATzilla}~\cite{xu_satzilla_2008} and \emph{AutoFolio}~\cite{lindauer-jair15a} but it is also well-known that these two systems use a lot more computation resources (several CPU days) than other systems;
  \item since the test and training data are published, the system will tend to overfit the scenarios if we will not \changedThree{regularly provide} new scenarios to reveal such overfitting.
\end{enumerate}

\paragraph{ICON Challenge on Algorithm Selection}

The ICON Challenge on Algorithm
Selection\footnote{\url{http://challenge.icon-fet.eu/}} provided a 
comparative evaluation of state-of-the-art algorithm selection systems.
The winner of the challenge was the zilla system~\cite{xu2012satzilla2012}.
In this competition, the algorithm selectors needed to be submitted \changedThree{before} a fixed deadline and
each system was executed on the organizers' hardware with some limitations (e.g., at most $12$ CPU hours for training on one scenario).
Although the used scenarios were also already published before (i.e., ASlib
version 1.0.1),
the organizers did not reveal the training-test splits to avoid \changedThree{overly} strong overfitting on these scenarios.
Furthermore, the ICON challenge assessed the performance of the algorithm selectors based on $3$ different performance metrics (i.e., average number of solved instances, PAR10, and misclassification penalty (MCP)) which revealed some strengths and weaknesses of algorithm selectors,
e.g., systems that used an algorithm schedule had better performance on solved instances and PAR10, but wasted some \changedThree{time with respect to} MCP.

\section{Summary}
\label{sec:conclusion}

We have introduced ASlib, a benchmark library for algorithm selection, a rapidly
growing field of research with substantial impact on various sub-communities in
artificial intelligence. 
ASlib facilitates research on algorithm selection methods by 
providing a common set of benchmarks and tools for working with these. Similar to solver competitions, 
it enables principled comparative empirical performance assessment.
It also considerably lowers the otherwise rather high barrier for researchers to work on algorithm selection,
since anyone using the benchmark scenarios we provide does not have to perform
actual runs of the solvers contained in them.
Since our library provides performance data for the solvers and problem instances
included in each selection scenario, using ASlib also substantially reduces the computational burden of performance assessments. \changedThree{Otherwise, these data would have to be produced, at considerable computational cost, by anyone
working with that scenario.} We carefully selected the set of scenarios included in release version 2.0 of
ASlib to challenge algorithm selection 
methods in various ways and thus provide a solid basis for developing and assessing such methods.

Release version 2.0 of the library contains 17 algorithm selection
scenarios from six different areas with a focus on (but not a limitation to) constraint satisfaction problems.
We discussed the format of new algorithm selection scenarios and showed examples
of the automated exploratory data analysis that will run for each new scenario
submitted to our online platform 
\url{http://aslib.net/}.
Finally, exploratory studies with various algorithm selection approaches
demonstrated the performance \changedThree{that} algorithm selection systems can achieve on our
scenarios.

\subsection*{Acknowledgements}
\label{sec:ack}


We thank the creators of the algorithms and instance distributions used in our various algorithm selection scenarios. 
The performance of algorithm selection systems depends critically upon the ingenuity and tireless efforts of domain experts who continue to invent novel solver strategies.

FH and ML are supported by the DFG (German Research Foundation) under Emmy
Noether grant HU 1900/2-1. KLB, AF and LK were supported by an NSERC E.W.R.\
Steacie Fellowship; in addition, all of these, along with HH, were supported
under the NSERC Discovery Grant Program. Part of this research was supported by
an Azure for Research grant.

\bibliographystyle{elsarticle-harv}
\bibliography{local}

\end{document}


\begin{table}[ht]
\centering
\begin{tabular}{lrrr}
  \toprule
Model & Solved & PAR10 & MCP \\ 
  \midrule
baseline/singleBestBySuccesses & 0.86 & 880.55 & 58.88 \\ 
  baseline/singleBest & 0.86 & 880.55 & 58.88 \\ 
  baseline/singleBestByPar & 0.86 & 880.55 & 58.88 \\ 
  baseline/vbs & 0.94 & 400.18 & 0.00 \\ 
  classif/ksvm & 0.89 & 693.12 & 33.14 \\ 
  classif/randomForest & 0.89 & 674.88 & 27.46 \\ 
  classif/rpart & 0.89 & 654.97 & 28.43 \\ 
  cluster/XMeans & 0.89 & 688.78 & 37.14 \\ 
  regr/lm & 0.90 & 615.05 & 26.04 \\ 
  regr/randomForest & 0.92 & 495.11 & 14.56 \\ 
  regr/rpart & 0.90 & 625.06 & 31.85 \\ 
   \bottomrule
\end{tabular}
\caption{Full experimental results for scenario ASP-POTASSCO. The following feature steps were used: Static (38 features).
The cost for using the feature steps (adapted for presolving) is 1404.25, on average 1.0852 per instance.} 
\end{table}
\begin{table}[ht]
\centering
\begin{tabular}{lrrr}
  \toprule
Model & Solved & PAR10 & MCP \\ 
  \midrule
baseline/singleBestBySuccesses & 0.86 & 7201.56 & 79.14 \\ 
  baseline/singleBest & 0.86 & 7201.56 & 79.14 \\ 
  baseline/singleBestByPar & 0.86 & 7201.56 & 79.14 \\ 
  baseline/vbs & 0.88 & 6344.25 & 0.00 \\ 
  classif/ksvm & 0.87 & 6729.74 & 29.76 \\ 
  classif/randomForest & 0.87 & 6629.11 & 18.06 \\ 
  classif/rpart & 0.87 & 6488.16 & 10.51 \\ 
  cluster/XMeans & 0.86 & 7005.38 & 60.83 \\ 
  regr/lm & 0.87 & 6633.48 & 22.43 \\ 
  regr/randomForest & 0.87 & 6542.55 & 20.43 \\ 
  regr/rpart & 0.86 & 6952.25 & 52.17 \\ 
   \bottomrule
\end{tabular}
\caption{Full experimental results for scenario CSP-2010. The following feature steps were used: all-feats (86 features).
} 
\end{table}
\begin{table}[ht]
\centering
\begin{tabular}{lrrr}
  \toprule
Model & Solved & PAR10 & MCP \\ 
  \midrule
baseline/singleBestBySuccesses & 0.52 & 8789.17 & 492.88 \\ 
  baseline/singleBest & 0.52 & 8789.17 & 492.88 \\ 
  baseline/singleBestByPar & 0.52 & 8789.17 & 492.88 \\ 
  baseline/vbs & 0.80 & 3752.47 & 0.00 \\ 
  classif/ksvm & 0.75 & 4669.17 & 87.47 \\ 
  classif/randomForest & 0.74 & 4831.28 & 103.72 \\ 
  classif/rpart & 0.71 & 5355.83 & 143.84 \\ 
  cluster/XMeans & 0.56 & 7976.18 & 405.86 \\ 
  regr/lm & 0.75 & 4592.36 & 80.58 \\ 
  regr/randomForest & 0.77 & 4205.36 & 47.04 \\ 
  regr/rpart & 0.74 & 4908.38 & 115.47 \\ 
   \bottomrule
\end{tabular}
\caption{Full experimental results for scenario CSP-MZN-2013. The following feature steps were used: static (144 features).
The cost for using the feature steps (adapted for presolving) is 340947, on average 73.4483 per instance.} 
\end{table}
\begin{table}[ht]
\centering
\begin{tabular}{lrrr}
  \toprule
Model & Solved & PAR10 & MCP \\ 
  \midrule
baseline/singleBestBySuccesses & 0.77 & 4893.14 & 190.90 \\ 
  baseline/singleBest & 0.77 & 4893.14 & 190.90 \\ 
  baseline/singleBestByPar & 0.77 & 4893.14 & 190.90 \\ 
  baseline/vbs & 0.85 & 3127.24 & 0.00 \\ 
  classif/ksvm & 0.82 & 3789.76 & 58.31 \\ 
  classif/randomForest & 0.82 & 3745.49 & 57.19 \\ 
  classif/rpart & 0.82 & 3795.52 & 64.07 \\ 
  cluster/XMeans & 0.82 & 3833.90 & 80.87 \\ 
  regr/lm & 0.82 & 3777.87 & 68.00 \\ 
  regr/randomForest & 0.84 & 3297.74 & 19.37 \\ 
  regr/rpart & 0.82 & 3789.72 & 79.84 \\ 
   \bottomrule
\end{tabular}
\caption{Full experimental results for scenario MAXSAT12-PMS. The following feature steps were used: group-basics (37 features).
The cost for using the feature steps (adapted for presolving) is 132.54, on average 0.151301 per instance.} 
\end{table}
\begin{table}[ht]
\centering
\begin{tabular}{lrrr}
  \toprule
Model & Solved & PAR10 & MCP \\ 
  \midrule
baseline/singleBestBySuccesses & 0.88 & 2231.05 & 118.36 \\ 
  baseline/singleBest & 0.88 & 2231.05 & 118.36 \\ 
  baseline/singleBestByPar & 0.88 & 2231.05 & 118.36 \\ 
  baseline/vbs & 0.93 & 1357.95 & 0.00 \\ 
  classif/ksvm & 0.89 & 1957.04 & 59.99 \\ 
  classif/randomForest & 0.90 & 1905.59 & 62.45 \\ 
  classif/rpart & 0.89 & 1992.38 & 68.38 \\ 
  cluster/XMeans & 0.90 & 1839.30 & 77.02 \\ 
  regr/lm & 0.90 & 1830.39 & 68.12 \\ 
  regr/randomForest & 0.91 & 1669.72 & 42.23 \\ 
  regr/rpart & 0.89 & 2063.11 & 85.20 \\ 
   \bottomrule
\end{tabular}
\caption{Full experimental results for scenario MAXSAT15-PMS-INDU. The following feature steps were used: ALL (37 features).
} 
\end{table}
\begin{table}[ht]
\centering
\begin{tabular}{lrrr}
  \toprule
Model & Solved & PAR10 & MCP \\ 
  \midrule
baseline/singleBestBySuccesses & 0.81 & 7002.91 & 688.77 \\ 
  baseline/singleBest & 0.81 & 7002.91 & 688.77 \\ 
  baseline/singleBestByPar & 0.81 & 7002.91 & 688.77 \\ 
  baseline/vbs & 1.00 & 227.60 & 0.00 \\ 
  classif/ksvm & 0.86 & 5423.07 & 522.98 \\ 
  classif/randomForest & 0.84 & 5908.73 & 578.28 \\ 
  classif/rpart & 0.86 & 5437.73 & 537.64 \\ 
  cluster/XMeans & 0.85 & 5534.53 & 511.48 \\ 
  regr/lm & 0.84 & 5887.33 & 556.88 \\ 
  regr/randomForest & 0.87 & 5028.53 & 497.32 \\ 
  regr/rpart & 0.84 & 5916.12 & 585.67 \\ 
   \bottomrule
\end{tabular}
\caption{Full experimental results for scenario PREMARSHALLING-ASTAR-2015. The following feature steps were used: orig (16 features).
} 
\end{table}
\begin{table}[ht]
\centering
\begin{tabular}{lrrr}
  \toprule
Model & Solved & PAR10 & MCP \\ 
  \midrule
baseline/singleBestBySuccesses & 0.63 & 13443.35 & 957.46 \\ 
  baseline/singleBest & 0.63 & 13443.35 & 957.46 \\ 
  baseline/singleBestByPar & 0.63 & 13443.35 & 957.46 \\ 
  baseline/vbs & 0.89 & 4105.89 & 0.00 \\ 
  classif/ksvm & 0.82 & 6623.25 & 273.52 \\ 
  classif/randomForest & 0.82 & 6616.83 & 275.17 \\ 
  classif/rpart & 0.75 & 9019.04 & 509.89 \\ 
  cluster/XMeans & 0.77 & 8425.42 & 472.67 \\ 
  regr/lm & 0.83 & 6246.41 & 235.03 \\ 
  regr/randomForest & 0.84 & 5867.26 & 202.24 \\ 
  regr/rpart & 0.81 & 6848.25 & 296.95 \\ 
   \bottomrule
\end{tabular}
\caption{Full experimental results for scenario PROTEUS-2014. The following feature steps were used: csp (36 features).
The cost for using the feature steps (adapted for presolving) is 26059.7, on average 6.4809 per instance.} 
\end{table}
\begin{table}[ht]
\centering
\begin{tabular}{lrrr}
  \toprule
Model & Solved & PAR10 & MCP \\ 
  \midrule
baseline/singleBestBySuccesses & 0.58 & 15330.17 & 716.76 \\ 
  baseline/singleBest & 0.58 & 15330.17 & 716.76 \\ 
  baseline/singleBestByPar & 0.58 & 15330.17 & 716.76 \\ 
  baseline/vbs & 0.77 & 8337.10 & 0.00 \\ 
  classif/ksvm & 0.71 & 10665.59 & 244.28 \\ 
  classif/randomForest & 0.73 & 9828.03 & 164.61 \\ 
  classif/rpart & 0.72 & 10293.05 & 227.01 \\ 
  cluster/XMeans & 0.66 & 12343.24 & 429.82 \\ 
  regr/lm & 0.71 & 10388.71 & 227.93 \\ 
  regr/randomForest & 0.75 & 9177.62 & 106.31 \\ 
  regr/rpart & 0.71 & 10405.54 & 244.75 \\ 
   \bottomrule
\end{tabular}
\caption{Full experimental results for scenario QBF-2011. The following feature steps were used: all-feats (46 features).
} 
\end{table}
\begin{table}[ht]
\centering
\begin{tabular}{lrrr}
  \toprule
Model & Solved & PAR10 & MCP \\ 
  \midrule
baseline/singleBestBySuccesses & 0.63 & 3401.81 & 170.95 \\ 
  baseline/singleBest & 0.63 & 3401.81 & 170.95 \\ 
  baseline/singleBestByPar & 0.63 & 3401.81 & 170.95 \\ 
  baseline/vbs & 0.81 & 1758.14 & 0.00 \\ 
  classif/ksvm & 0.71 & 2653.52 & 87.96 \\ 
  classif/randomForest & 0.72 & 2529.65 & 73.91 \\ 
  classif/rpart & 0.69 & 2783.88 & 102.06 \\ 
  cluster/XMeans & 0.64 & 3290.49 & 156.51 \\ 
  regr/lm & 0.69 & 2824.38 & 110.26 \\ 
  regr/randomForest & 0.77 & 2104.42 & 36.24 \\ 
  regr/rpart & 0.72 & 2548.49 & 86.28 \\ 
   \bottomrule
\end{tabular}
\caption{Full experimental results for scenario QBF-2014. The following feature steps were used: default (46 features).
} 
\end{table}
\begin{table}[ht]
\centering
\begin{tabular}{lrrr}
  \toprule
Model & Solved & PAR10 & MCP \\ 
  \midrule
baseline/singleBestBySuccesses & 0.50 & 25589.27 & 1434.69 \\ 
  baseline/singleBest & 0.50 & 25649.09 & 1342.48 \\ 
  baseline/singleBestByPar & 0.50 & 25589.27 & 1434.69 \\ 
  baseline/vbs & 0.74 & 13360.66 & 0.00 \\ 
  classif/ksvm & 0.67 & 16795.52 & 368.56 \\ 
  classif/randomForest & 0.65 & 17960.50 & 470.43 \\ 
  classif/rpart & 0.64 & 18329.71 & 535.66 \\ 
  cluster/XMeans & 0.52 & 24067.25 & 958.47 \\ 
  regr/lm & 0.61 & 19715.73 & 554.42 \\ 
  regr/randomForest & 0.66 & 17595.72 & 408.96 \\ 
  regr/rpart & 0.61 & 19835.52 & 674.20 \\ 
   \bottomrule
\end{tabular}
\caption{Full experimental results for scenario SAT11-HAND. The following feature steps were used: Pre (6 features).
The cost for using the feature steps (adapted for presolving) is 290.88, on average 0.982703 per instance.} 
\end{table}
\begin{table}[ht]
\centering
\begin{tabular}{lrrr}
  \toprule
Model & Solved & PAR10 & MCP \\ 
  \midrule
baseline/singleBestBySuccesses & 0.72 & 14605.90 & 718.39 \\ 
  baseline/singleBest & 0.72 & 14605.90 & 718.39 \\ 
  baseline/singleBestByPar & 0.72 & 14605.90 & 718.39 \\ 
  baseline/vbs & 0.84 & 8187.52 & 0.00 \\ 
  classif/ksvm & 0.73 & 13963.88 & 577.90 \\ 
  classif/randomForest & 0.74 & 13661.59 & 568.83 \\ 
  classif/rpart & 0.73 & 13915.25 & 525.25 \\ 
  cluster/XMeans & 0.72 & 14537.56 & 692.64 \\ 
  regr/lm & 0.73 & 13756.72 & 526.94 \\ 
  regr/randomForest & 0.72 & 14545.27 & 559.57 \\ 
  regr/rpart & 0.68 & 16443.08 & 817.67 \\ 
   \bottomrule
\end{tabular}
\caption{Full experimental results for scenario SAT11-INDU. The following feature steps were used: Pre (6 features).
The cost for using the feature steps (adapted for presolving) is 3332, on average 11.1067 per instance.} 
\end{table}
\begin{table}[ht]
\centering
\begin{tabular}{lrrr}
  \toprule
Model & Solved & PAR10 & MCP \\ 
  \midrule
baseline/singleBestBySuccesses & 0.60 & 19916.36 & 979.92 \\ 
  baseline/singleBest & 0.60 & 19916.36 & 979.92 \\ 
  baseline/singleBestByPar & 0.60 & 19916.36 & 979.92 \\ 
  baseline/vbs & 0.82 & 9186.44 & 0.00 \\ 
  classif/ksvm & 0.77 & 11806.90 & 205.86 \\ 
  classif/randomForest & 0.79 & 10831.69 & 130.67 \\ 
  classif/rpart & 0.79 & 10644.77 & 93.24 \\ 
  cluster/XMeans & 0.72 & 14071.00 & 447.01 \\ 
  regr/lm & 0.78 & 10979.25 & 127.61 \\ 
  regr/randomForest & 0.80 & 10260.29 & 83.75 \\ 
  regr/rpart & 0.78 & 11103.93 & 177.72 \\ 
   \bottomrule
\end{tabular}
\caption{Full experimental results for scenario SAT11-RAND. The following feature steps were used: Pre (6 features).
The cost for using the feature steps (adapted for presolving) is 47.11, on average 0.0785167 per instance.} 
\end{table}
\begin{table}[ht]
\centering
\begin{tabular}{lrrr}
  \toprule
Model & Solved & PAR10 & MCP \\ 
  \midrule
baseline/singleBestBySuccesses & 0.75 & 3079.89 & 302.51 \\ 
  baseline/singleBest & 0.75 & 3079.89 & 302.51 \\ 
  baseline/singleBestByPar & 0.75 & 3079.89 & 302.51 \\ 
  baseline/vbs & 0.99 & 241.32 & 0.00 \\ 
  classif/ksvm & 0.90 & 1286.27 & 97.15 \\ 
  classif/randomForest & 0.92 & 1083.40 & 80.91 \\ 
  classif/rpart & 0.77 & 2897.47 & 241.13 \\ 
  cluster/XMeans & 0.75 & 3115.03 & 296.30 \\ 
  regr/lm & 0.86 & 1851.19 & 154.37 \\ 
  regr/randomForest & 0.92 & 1044.39 & 74.47 \\ 
  regr/rpart & 0.83 & 2160.57 & 191.10 \\ 
   \bottomrule
\end{tabular}
\caption{Full experimental results for scenario SAT12-ALL. The following feature steps were used: Pre (6 features).
The cost for using the feature steps (adapted for presolving) is 9021.19, on average 5.58934 per instance.} 
\end{table}
\begin{table}[ht]
\centering
\begin{tabular}{lrrr}
  \toprule
Model & Solved & PAR10 & MCP \\ 
  \midrule
baseline/singleBestBySuccesses & 0.48 & 6338.90 & 254.76 \\ 
  baseline/singleBest & 0.48 & 6338.90 & 254.76 \\ 
  baseline/singleBestByPar & 0.48 & 6338.90 & 254.76 \\ 
  baseline/vbs & 0.70 & 3662.24 & 0.00 \\ 
  classif/ksvm & 0.61 & 4738.13 & 95.27 \\ 
  classif/randomForest & 0.63 & 4462.59 & 72.81 \\ 
  classif/rpart & 0.56 & 5349.11 & 143.27 \\ 
  cluster/XMeans & 0.50 & 6118.22 & 223.52 \\ 
  regr/lm & 0.59 & 4961.73 & 107.51 \\ 
  regr/randomForest & 0.65 & 4320.67 & 57.29 \\ 
  regr/rpart & 0.58 & 5140.14 & 131.32 \\ 
   \bottomrule
\end{tabular}
\caption{Full experimental results for scenario SAT12-HAND. The following feature steps were used: Pre (6 features).
The cost for using the feature steps (adapted for presolving) is 801.95, on average 1.04557 per instance.} 
\end{table}
\begin{table}[ht]
\centering
\begin{tabular}{lrrr}
  \toprule
Model & Solved & PAR10 & MCP \\ 
  \midrule
baseline/singleBestBySuccesses & 0.74 & 3266.05 & 128.35 \\ 
  baseline/singleBest & 0.74 & 3266.05 & 128.35 \\ 
  baseline/singleBestByPar & 0.74 & 3266.05 & 128.35 \\ 
  baseline/vbs & 0.82 & 2221.50 & 0.00 \\ 
  classif/ksvm & 0.77 & 2875.11 & 55.50 \\ 
  classif/randomForest & 0.78 & 2806.37 & 41.43 \\ 
  classif/rpart & 0.72 & 3449.57 & 121.24 \\ 
  cluster/XMeans & 0.72 & 3433.01 & 118.63 \\ 
  regr/lm & 0.76 & 3015.79 & 75.81 \\ 
  regr/randomForest & 0.78 & 2770.11 & 42.38 \\ 
  regr/rpart & 0.74 & 3272.58 & 105.62 \\ 
   \bottomrule
\end{tabular}
\caption{Full experimental results for scenario SAT12-INDU. The following feature steps were used: Pre (6 features).
The cost for using the feature steps (adapted for presolving) is 22200.2, on average 19.0234 per instance.} 
\end{table}
\begin{table}[ht]
\centering
\begin{tabular}{lrrr}
  \toprule
Model & Solved & PAR10 & MCP \\ 
  \midrule
baseline/singleBestBySuccesses & 0.73 & 3271.14 & 49.40 \\ 
  baseline/singleBest & 0.73 & 3271.14 & 49.40 \\ 
  baseline/singleBestByPar & 0.73 & 3271.14 & 49.40 \\ 
  baseline/vbs & 0.76 & 2872.84 & 0.00 \\ 
  classif/ksvm & 0.73 & 3272.16 & 44.50 \\ 
  classif/randomForest & 0.73 & 3231.00 & 42.96 \\ 
  classif/rpart & 0.74 & 3133.81 & 32.82 \\ 
  cluster/XMeans & 0.73 & 3231.30 & 43.25 \\ 
  regr/lm & 0.74 & 3161.09 & 36.31 \\ 
  regr/randomForest & 0.74 & 3169.67 & 37.00 \\ 
  regr/rpart & 0.72 & 3471.65 & 69.66 \\ 
   \bottomrule
\end{tabular}
\caption{Full experimental results for scenario SAT12-RAND. The following feature steps were used: Pre (6 features).
The cost for using the feature steps (adapted for presolving) is 80.14, on average 0.0588399 per instance.} 
\end{table}
\begin{table}[ht]
\centering
\begin{tabular}{lrrr}
  \toprule
Model & Solved & PAR10 & MCP \\ 
  \midrule
baseline/singleBestBySuccesses & 0.87 & 5189.36 & 525.79 \\ 
  baseline/singleBest & 0.87 & 5189.36 & 525.79 \\ 
  baseline/singleBestByPar & 0.87 & 5189.36 & 525.79 \\ 
  baseline/vbs & 0.94 & 2287.57 & 0.00 \\ 
  classif/ksvm & 0.89 & 4458.57 & 335.00 \\ 
  classif/randomForest & 0.88 & 4562.34 & 330.77 \\ 
  classif/rpart & 0.88 & 4587.28 & 355.71 \\ 
  cluster/XMeans & 0.86 & 5483.98 & 496.41 \\ 
  regr/lm & 0.87 & 4930.56 & 374.99 \\ 
  regr/randomForest & 0.91 & 3763.44 & 287.87 \\ 
  regr/rpart & 0.87 & 5119.23 & 455.66 \\ 
   \bottomrule
\end{tabular}
\caption{Full experimental results for scenario SAT15-INDU. The following feature steps were used: ALL (54 features).
} 
\end{table}